\pdfoutput=1

\documentclass[11pt]{article}

\usepackage[preprint]{acl}

\usepackage{times}
\usepackage{latexsym}

\usepackage[T1]{fontenc}

\usepackage[utf8]{inputenc}

\usepackage{microtype}

\usepackage{inconsolata}

\usepackage{graphicx}
\usepackage{hyperref}
\usepackage{url}
\usepackage{enumitem}
\usepackage[most]{tcolorbox}
\newcommand{\xhdr}[1]{{\noindent\bfseries #1}.}
\usepackage{booktabs}
\usepackage{booktabs}
\usepackage{colortbl}
\usepackage{caption}
\usepackage{wrapfig}
\usepackage{subfigure}
\usepackage{CJKutf8}
\usepackage{multirow}
\definecolor{mypink}{rgb}{.99,.91,.95}
\definecolor{mygreen}{rgb}{.9,.99,.9}
\definecolor{mygray}{gray}{.9}
\definecolor{Blue}{rgb}{0.05,0.05,0.4}
\definecolor{Red}{rgb}{0.4,0.05,0.05}
\newcommand{\bluenote}[1]{\textcolor{Blue}{#1}}
\newcommand{\rednote}[1]{\textcolor{Red}{#1}}

\title{LongBench v2: Towards Deeper Understanding and\\ Reasoning on Realistic Long-context Multitasks}

\author{Yushi Bai$^{1*}$, Shangqing Tu$^{1*}$, Jiajie Zhang$^1$, Hao Peng$^1$, \\
\textbf{Xiaozhi Wang$^1$, Xin Lv$^2$, Shulin Cao$^2$, Jiazheng Xu$^1$,} \\
\textbf{Lei Hou$^1$, Yuxiao Dong$^1$, Jie Tang$^1$, Juanzi Li$^1$} \\
$^1$Tsinghua University
  \quad
$^2$Zhipu.AI \\
\url{https://longbench2.github.io}
\phantom{\thanks{Equal contribution. Author contributions are listed in Appendix~\ref{sec:ac}.}}
  }

\begin{document}
\maketitle


\begin{abstract}

This paper introduces LongBench v2, a benchmark designed to assess the ability of LLMs to handle long-context problems requiring \emph{deep understanding and reasoning} across real-world multitasks. LongBench v2 consists of 503 challenging multiple-choice questions, with contexts ranging from 8k to 2M words, across six major task categories: single-document QA, multi-document QA, long in-context learning, long-dialogue history understanding, code repository understanding, and long structured data understanding.
To ensure the breadth and the practicality, we collect data from nearly 100 highly educated individuals with diverse professional backgrounds. We employ both automated and manual review processes to maintain high quality and difficulty, resulting in human experts achieving only 53.7\% accuracy under a 15-minute time constraint.
Our evaluation reveals that the best-performing model, when directly answers the questions, achieves only 50.1\% accuracy. In contrast, the o1-preview model, which includes longer reasoning, achieves 57.7\%, surpassing the human baseline by 4\%. 
These results highlight the importance of enhanced reasoning ability and scaling inference-time compute to tackle the long-context challenges in LongBench v2.

\end{abstract}

\section{Introduction}

Over the past year, research and products on long-context large language models (LLMs) have made remarkable progress: in terms of context window length, advancing from the initial 8k to the current 128k and even 1M tokens~\cite{GPT-4o,claude-3-5,reid2024gemini,glm2024chatglm}; and achieving promising performance on long-context benchmarks. However, beneath these advancements lies an urgent and practical question: \textbf{Do these models truly comprehend the long texts they process, i.e., are they capable of deeply understanding, learning, and reasoning based on the information contained in these long texts?}

Critically, existing long-context understanding benchmarks~\cite{bai2024longbench,zhang2024infty,hsieh2024ruler} fail to reflect the long-context LLMs' \emph{deep} understanding capabilities across diverse tasks.
They often focus on extractive questions, where answers are directly found in the material, a challenge easily handled by modern long-context models and RAG systems, as evidenced by their perfect recall in the Needle-in-a-Haystack test~\cite{needleinhaystack}.
Furthermore, many of these benchmarks rely on synthetic tasks, which limits their applicability to real-world scenarios, and their adopted metrics like F1 and ROUGE are unreliable.

To address these issues, we aim to build a benchmark with the following features: 
(1) \textbf{Length}: Context length ranging from 8k to 2M words, with the majority under 128k.
(2) \textbf{Difficulty}: Challenging enough that even human experts, using search tools within the document, cannot answer correctly in a short time.
(3) \textbf{Coverage}: Cover various realistic scenarios.
(4) \textbf{Reliability}: All in a multiple-choice question format for reliable evaluation.

\begin{figure}[t]
    \centering
    \includegraphics[width=\linewidth]{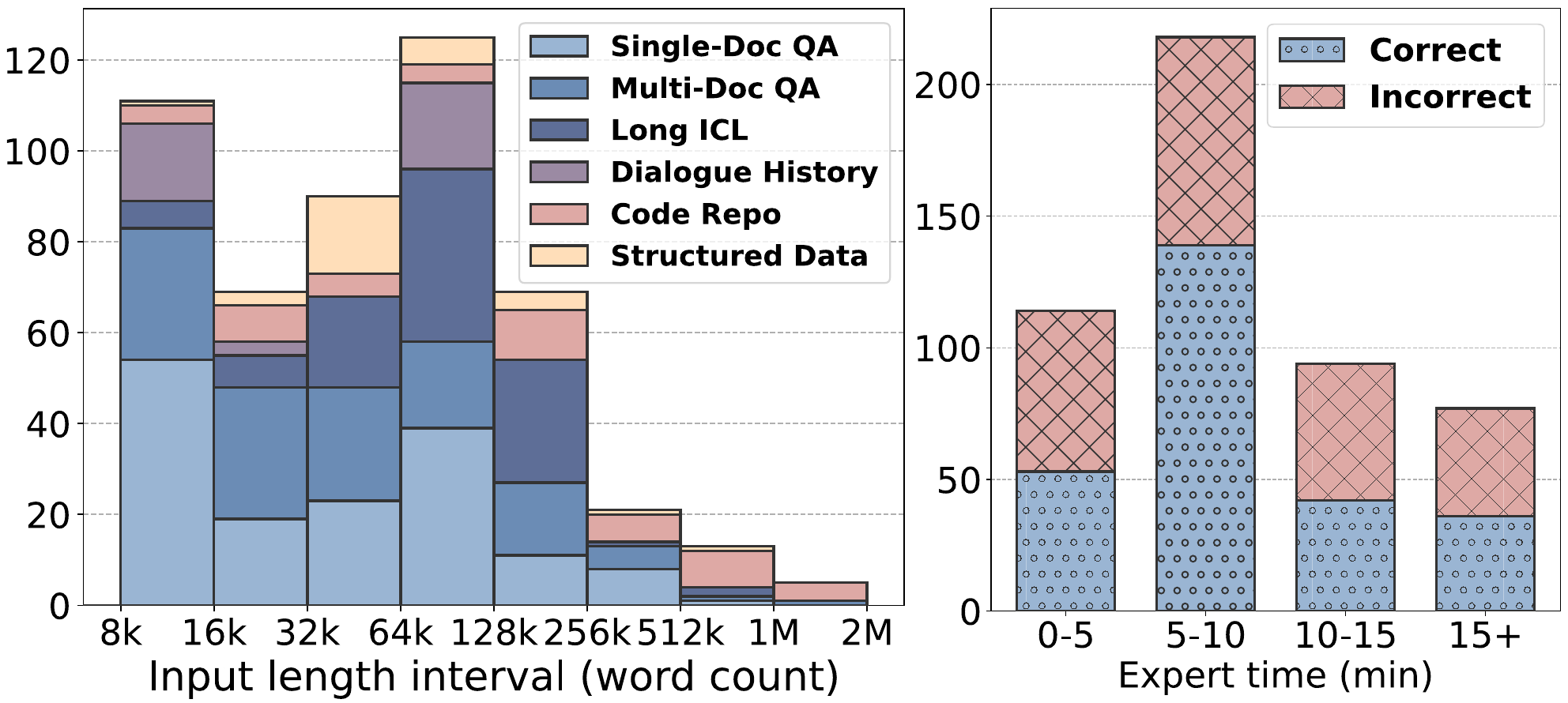}
    \caption{Length distribution (left) and human expert solving time distribution (right) of LongBench v2.}
    \label{fig:length}
\end{figure}

With the above goal in mind, we present \emph{LongBench v2}.
LongBench v2 contains 503 multiple-choice questions and is made up of 6 major task categories and 20 subtasks to cover as many realistic deep comprehension scenarios as possible, including \emph{single-document QA}, \emph{multi-document QA}, \emph{long in-context learning}, \emph{long-dialogue history understanding}, \emph{code repository understanding}, and \emph{long structured data understanding} (detailed in Table~\ref{tb:stat}).
All the test data in LongBench v2 are in English, and the length distribution of each task category is shown on the left of Figure~\ref{fig:length}.

To ensure the quality and difficulty of test data, we combine automated and manual reviews during data collection. 
We first recruit 97 data annotators with diverse academic backgrounds and grades from top universities and then select 24 data reviewers from this group.
Annotators provide data including long documents, questions, options, answers, and evidence.
We then leverage three long-context LLMs for an automated review, where a question is considered too easy if all three LLMs answer it correctly.
Data passing the automated review are assigned to the reviewers, who answer the questions and determine whether the questions are appropriate (meet our requirements) and if the answers are correct.
In our criteria, a qualified data point should have (1) an appropriate question with an objective, correct answer; (2) sufficient difficulty, such that all three LLMs cannot answer correctly at the same time, and the human reviewer cannot answer correctly within 3 minutes, even with searching tools within the document.
If data do not meet these criteria, we request modifications from the annotator.
We also set length and difficulty incentives to encourage longer and harder test data.
Figure~\ref{fig:length} (right) visualizes the distribution of expert solving times along with human accuracy.

Overall, our data shows a median word count of 54k and an average of 104k words. 
Human experts are able to achieve an accuracy of only 53.7\% within 15 minutes, compared to 25\% accuracy with random guessing, highlighting the challenging nature of the test.
In the evaluation, the best-performing model achieves only 50.1\% accuracy when directly outputting the answer. In contrast, the o1-preview model, which incorporates longer reasoning during inference, reaches 57.7\%, surpassing human experts. This implies that LongBench v2 places greater demands on the reasoning ability of current models, and incorporating more inference-time thinking and reasoning appears to be a natural and crucial step in addressing such long-context reasoning challenges.
We hope LongBench v2 will accelerate the exploration of how scaling inference-time compute will affect deep understanding and reasoning in long-context scenarios.

\section{Related Work}

We divide existing long-context benchmarks for LLMs into two types. 
The first consists of comprehensive benchmarks that combine multitasks such as QA, retrieval, and summarization. 
Sorted by publication date, these benchmarks include ZeroSCROLLS~\cite{shaham2023zeroscrolls}, L-Eval~\cite{an2024leval}, LongBench~\cite{bai2024longbench}, BAMBOO~\cite{dong2024bamboo}, LooGLE~\cite{li2023loogle}, $\infty$-bench~\cite{zhang2024infty}, Ruler~\cite{hsieh2024ruler}, and HELMET~\cite{yen2024helmet}.
It is noteworthy that most of these multitask benchmarks were proposed last year, which corresponds to the thrive of long-context LLMs, whose context length has been extended to 128k tokens or more~\cite{claude-3-5,GPT-4o,reid2024gemini,glm2024chatglm,dubey2024llama} through continual training~\cite{xiong2024effective,pmlr-v235-fu24d,bai2024longalign,gao2024train}.

The other category of long-context benchmarks is more targeted, evaluating models on specific types of long-context tasks, including document QA~\cite{kovcisky2018narrativeqa,dua2019drop,dasigi2021dataset,pang2022quality,wang2024leave}, summarization~\cite{zhong2021qmsum,huang2021efficient,wang2022squality}, retrieval and attributing~\cite{needleinhaystack,kuratov2024babilong,song2024counting,laban2024summary,zhang2024longcite,vodrahalli2024michelangelo,krishna2024fact}, conversation~\cite{bai2024longalign}, coding~\cite{liu2023repobench,bogomolov2024long}, many-shot learning~\cite{agarwal2024many}, and long-text generation~\cite{bai2024longwriter,wu2024longgenbench,liu2024longgenbench,que2024hellobench}.

In our view, existing long-context benchmarks generally have the following issues: (1) \emph{Lack of deep reasoning}: While a few benchmarks contain longer examples of around 100k, most of these data have not been human-examined, and many of these samples can be solved through shallow understanding such as retrieval, thus failing to reflect a model's deep reasoning capabilities.
(2) \emph{Unreliable metrics}: Many datasets use metrics like ROUGE and F1 for evaluation, which are known to be unreliable~\cite{novikova2017we}. Additionally, some datasets adopt LLM-as-a-judge~\cite{zheng2023judging,li2024generation} for evaluation, which can be costly and may introduce biases in their assessments~\cite{bai2024benchmarking,ye2024justice}.
To construct a more challenging, reliable, and comprehensive long-context benchmark, we employ a uniform multiple-choice format and manually verify each data point to ensure it meets the required level of difficulty.

\begin{table*}[t]
\centering  
\resizebox{\textwidth}{!}{
\begin{tabular}{llrrrr}
\toprule
\textbf{Dataset} & \textbf{Source} & \textbf{\#data} & \textbf{Length} & \textbf{Expert Acc} & \textbf{Expert Time$^*$} \\
\midrule
\multicolumn{2}{l}{\cellcolor{mypink}\emph{I. Single-Document QA}} & \cellcolor{mypink}175 & \cellcolor{mypink}51k & \cellcolor{mypink}55\% & \cellcolor{mypink}8.9 min \\
Academic & Paper, textbook & 44 & 14k & 50\% & 7.3 min \\
Literary & Novel & 30 & 72k & 47\% & 8.5 min \\
Legal & Legal doc & 19 & 15k & 53\% & 13.1 min \\
Financial & Financial report & 22 & 49k & 59\% & 9.0 min \\
Governmental & Government report & 18 & 20k & 50\% & 9.5 min \\
Detective & Detective novel & 22 & 70k & 64\% & 9.3 min \\
Event ordering & Novel & 20 & 96k & 75\% & 9.4 min \\
\midrule
\multicolumn{2}{l}{\cellcolor{mypink}\emph{II. Multi-Document QA}} & \cellcolor{mypink}125 & \cellcolor{mypink}34k & \cellcolor{mypink}36\% & \cellcolor{mypink}6.1 min \\
Academic & Papers, textbooks & 50 & 27k & 22\% & 6.1 min \\
Legal & Legal docs & 14 & 28k & 64\% & 8.8 min \\
Financial & Financial reports & 15 & 129k & 40\% & 7.0 min \\
Governmental & Government reports & 23 & 89k & 22\% & 6.0 min \\
Multi-news & News & 23 & 15k & 61\% & 5.3 min \\
\midrule
\multicolumn{2}{l}{\cellcolor{mypink}\emph{III. Long In-context Learning}} & \cellcolor{mypink}81 & \cellcolor{mypink}71k & \cellcolor{mypink}63\% & \cellcolor{mypink}8.3 min \\
User guide QA & Electronic device, software, instrument & 40 & 61k & 63\% & 9.9 min \\
New language translation & Vocabulary book (\textit{Kalamang}, \textit{Zhuang}) & 20 & 132k & 75\% & 5.4 min \\
Many-shot learning & Multi-class classification task & 21 & 71k & 52\% & 8.0 min \\
\midrule
\multicolumn{2}{l}{\cellcolor{mypink}\emph{IV. Long-dialogue History Understanding}} & \cellcolor{mypink}39 & \cellcolor{mypink}25k & \cellcolor{mypink}79\% & \cellcolor{mypink}8.2 min \\
Agent history QA & LLM agents conversation & 20 & 13k & 70\% & 8.3 min \\
Dialogue history QA & User-LLM conversation & 19 & 77k & 89\% & 6.5 min \\
\midrule
\multicolumn{2}{l}{\cellcolor{mypink}\emph{V. Code Repository Understanding}} & \cellcolor{mypink}50 & \cellcolor{mypink}167k & \cellcolor{mypink}44\% & \cellcolor{mypink}6.4 min \\
Code repo QA & Code repository & 50 & 167k & 44\% & 6.4 min \\
\midrule
\multicolumn{2}{l}{\cellcolor{mypink}\emph{VI. Long Structured Data Understanding}} & \cellcolor{mypink}33 & \cellcolor{mypink}49k & \cellcolor{mypink}73\% & \cellcolor{mypink}6.4 min \\
Table QA & Table & 18 & 42k & 61\% & 7.4 min \\
Knowledge graph reasoning & KG subgraph & 15 & 52k & 87\% & 6.2 min \\
\bottomrule
\end{tabular}
}
\caption{Tasks and data statistics in LongBench v2. `Source' denotes the origin of the context. `Length' is the \emph{median} of the number of words. `Expert Acc' and `Expert Time' refer to the average accuracy and the \emph{median} time spent on answering the question by human experts. 
$^*$: We allow human experts to respond with ``I don't know the answer'' if it takes them more than 15 minutes.
As a result, most expert times are under 15 minutes, but this doesn't necessarily mean that the questions are fully answered within such a time.}
\label{tb:stat}
\end{table*}

\section{LongBench v2: Task and Construction}

Our design principle focuses on four aspects: (1) The context should be sufficiently long to cover scenarios ranging from 8k to 2M words, with a relatively even distribution across texts up to 128k words.
(2) The question should be challenging, requiring the model to deeply understand the context to answer. It should avoid questions that can be answered based on memory or those where the answer can be directly extracted from the context.
(3) The data should cover a wide range of real-world long-context scenarios and reflect the model's holistic ability to reason, apply, and analyze information drawn from the lengthy text.
(4) The data should be in English and in a multiple-choice question format, containing a long text, a question, four choices, a groundtruth answer, and an evidence. Distractors should be included to prevent the model from guessing the correct answer based on option patterns.

\subsection{Task Overview}
Based on the testing scenarios and the types and sources of long texts, we propose six major task categories and further divide them into 20 subtasks.
We introduce the tasks included in LongBench v2 in the following. A list of task statistics and detailed descriptions can be found in Table~\ref{tb:stat} and Appendix~\ref{sec:task}.

\xhdr{Single-Doc QA}
We integrate subtask categories from previous datasets~\cite{bai2024longbench,an2024leval} and expand them to include QA for \emph{academic}, \emph{literary}, \emph{legal}, \emph{financial}, and \emph{governmental} documents. 
Considering that \emph{detective} QA~\cite{xu2024detectiveqa} requires in-depth reasoning based on case background, we introduce such a task that requires identifying the killer or motive based on information provided in detective novels.
We also include \emph{Event ordering}, where the goal is to order minor events according to the timeline of a novel.

\xhdr{Multi-Doc QA}
To distinguish from single-doc QA, multi-doc QA requires answers drawn from multiple provided documents.
Besides the categories in single-doc QA, multi-doc QA also includes \emph{multi-news QA}, which involves reasoning across multiple news articles, events, and timelines.

\begin{figure*}[t]
    \centering
    \includegraphics[width=\linewidth]{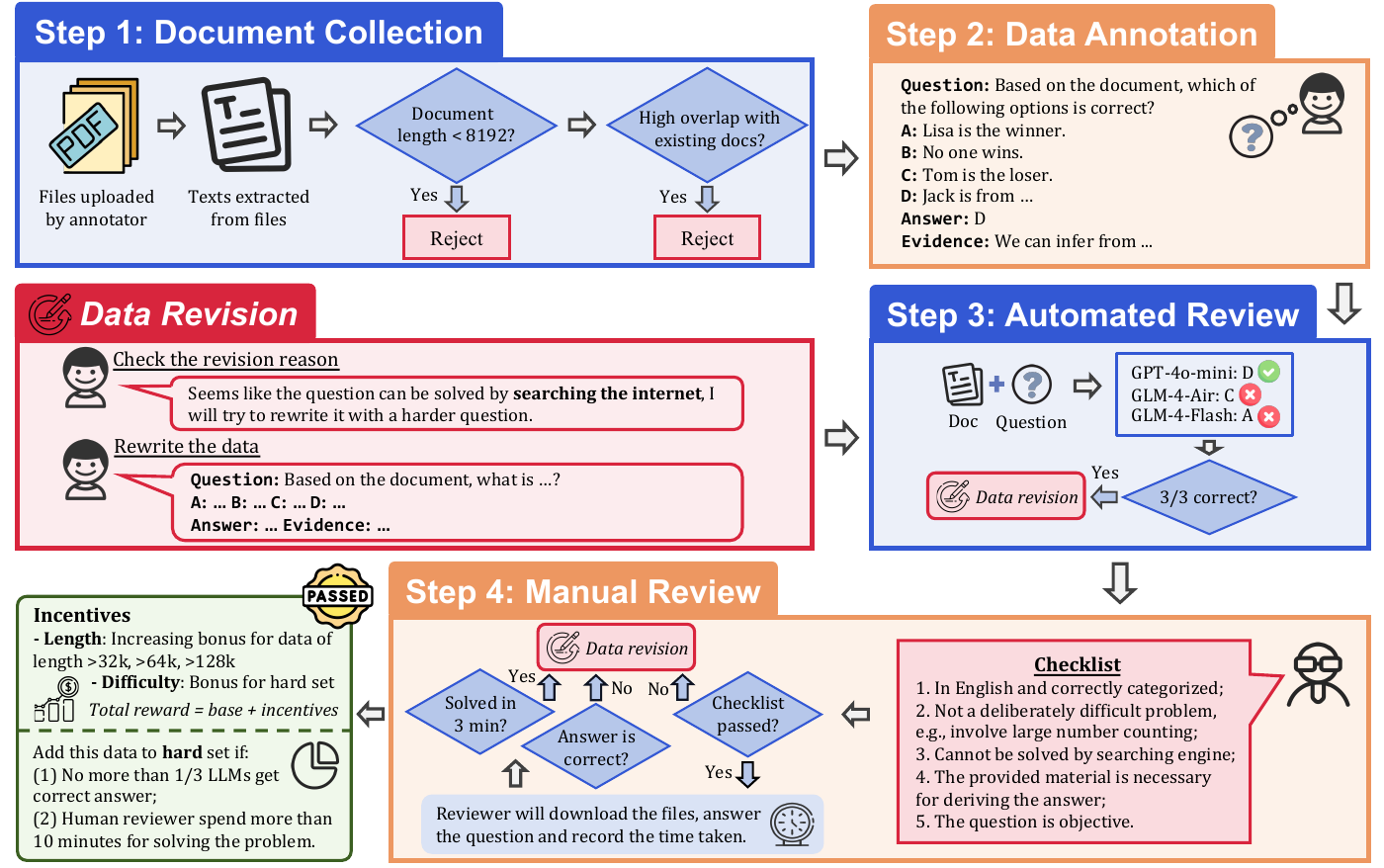}
    \caption{Data collection pipeline of LongBench v2. The annotator first uploads the document(s) and proposes a multiple-choice question based on the content. After that, automated and manual reviews will be conducted to ensure the data meets our requirements. Only data that passes these reviews is eligible for annotation rewards, meaning the annotator must revise the data until it passes all review stages. More details are in section~\ref{sec:data_collection}.}
    \label{fig:pipeline}
\end{figure*}

\xhdr{Long In-context Learning}
Learning from a long context, such as acquiring new skills, requires the ability to comprehend and reason based on that context. Hence, we consider it as a major category of tasks.
LongBench v2 includes several key tasks, including \emph{User guide QA}, which answers questions with information learnt from user guides for electronic devices, software, etc.; \emph{New language translation}~\cite{tanzerbenchmark,zhang2024teaching}, which involves learning to translate an unseen language from a vocabulary book; \emph{Many-shot learning}~\cite{agarwal2024many}, which involves learning to label new data from a handful of examples.

\xhdr{Long-dialogue History Understanding}
LLMs, as more intelligent chatbots or agents, require enhanced memory capabilities to handle longer histories. Therefore, we integrate long-dialogue history understanding tasks to test whether LLMs can handle information from long conversation histories.
These tasks are divided into two subtasks based on the source of the conversation history: one involving the history of interactions between multiple LLM agents, i.e., \emph{Agent history QA}~\cite{huang2024far}, and the other involving the dialogue history between a user and an LLM acting as an assistant, i.e., \emph{Dialogue history QA}~\cite{wu2024longmemeval}.

\xhdr{Code Repository Understanding}
Code repository contains long code content, and question answering over a code repository requires understanding and reasoning across multiple files, making it a common yet challenging long-context task.

\xhdr{Long Structured Data Understanding}
In addition to textual data, much information is presented in structured forms, so we introduce the long structured data QA task to test the LLM's understanding of long structured data, including reasoning on long tables, i.e., \emph{Table QA}~\cite{zhang2024tablellm}, and answering complex queries on knowledge graphs (KGs), i.e., \emph{Knowledge graph reasoning}~\cite{cao2022kqa,bai2023answering}.
We anonymize the entities in the KG to prevent the model from directly deriving the answers through memorization.

\subsection{Data Collection}
\label{sec:data_collection}

To collect high-quality and challenging data for long-context tasks, we hire 97 annotators who are either holding or pursuing a bachelor's degree from top universities and are proficient in English, with detailed statistics shown in Appendix~\ref{sec:stat}. We also select 24 professional human experts based on their major and year of study for conducting manual reviews.
Figure~\ref{fig:pipeline} illustrates the overall pipeline of our data collection process, which consists of five steps: document collection, data annotation, automated review, manual review, and data revision (optional). 
We develop an online annotation platform to implement this pipeline, with further details provided in Appendix~\ref{sec:platform}.

\xhdr{Step 1: Document Collection}
Unlike previous benchmarks~\cite{bai2024longbench,an2024leval}, where long documents are pre-defined or synthesized by the benchmark designers, we aim to gather documents that reflect more diverse scenarios and are more likely to be used in everyday contexts. To achieve this, we ask annotators to upload one or multiple files they have personally read or used, such as research papers, textbooks, novels, etc., according to the task type.
Our platform first converts the uploaded files into plain text using tools such as \href{https://github.com/pymupdf/PyMuPDF}{\texttt{PyMuPDF}}.
The input documents then undergo two automatic checks. If the length is less than 8,192 words, it is rejected as too short. Documents with a high overlap with previous annotations are also rejected to ensure diversity.

\xhdr{Step 2: Data Annotation}
During data annotation, the annotator is tasked with proposing a multiple-choice question based on their submitted documents. The question should be accompanied with four choices, a groundtruth answer, and the supporting evidence. We provide the annotators with a detailed question design principle that specifies our requirement (Appendix~\ref{sec:guide}). To summarize, the following types of questions should be avoided:
(1) \emph{Counting questions}: Avoid questions that require counting large numbers.
(2) \emph{Simple retrieval questions}: Do not ask basic information retrieval questions, as these are too easy for modern LLMs~\cite{song2024counting}.
(3) \emph{Overly professional questions}: Questions should not demand extensive external knowledge; they should rely on minimal expertise.
(4) \emph{Tricky questions}: Do not create questions that are deliberately difficult; the goal is to keep the questions natural and straightforward.

\xhdr{Step 3: Automated Review}
Upon submission, each question undergoes an initial automated review process to ensure it is not too easy. 
We employ three fast and powerful LLMs with a 128k context length to answer the questions: GPT-4o-mini~\cite{GPT-4o-mini}, \href{https://open.bigmodel.cn/pricing}{GLM-4-Air}, and \href{https://open.bigmodel.cn/pricing}{GLM-4-Flash}.
Inputs that exceed the context length are truncated from the middle.
If all three LLMs answer the question correctly, it is considered too easy. In such cases, annotators will be required to revise the question and choices to increase its difficulty.

\xhdr{Step 4: Manual Review}
Data passing the automated review is sent to a human expert for manual review. 
Our manual review serves two purposes: first, to filter out unqualified questions and data with incorrect answers; second, to establish a human baseline while also determining the difficulty of the questions and filter out those that are too easy (i.e., questions that humans can answer correctly in a short amount of time).
In practice, the reviewer first goes through a checklist to determine whether the question meets the specified requirements (outlined in Appendix~\ref{sec:guide}). Next, the reviewer downloads the raw document files and attempts to answer the question. The reviewer is encouraged to use searching tools within the files to solve the problem more promptly. Once a choice is submitted, the reviewer can view the groundtruth answer and the evidence provided by the annotators. The reviewer will then decide whether the answer is objective and fully correct. Our platform tracks the time spent on each question, and if the human expert answers correctly within 3 minutes, the question will be considered too easy, demanding a revision from its annotator.
Since answering some questions may require spending several hours reading the material, which implies a significant review time cost, we allow human experts to respond with ``I don't know the answer'' after 15 minutes.

\xhdr{Data Revision}
As mentioned above, questions deemed unqualified during either automated or manual review will require revision by its annotator. We set up a separate page in our platform for annotator to track their rejected data. For each rejected data, we provide the annotator with a reason for the rejection, classified into three categories: (1) \emph{Illegal question}: Rejected by human reviewers due to the question being unqualified, (2) \emph{Insufficient difficulty}: Rejected by automated review or due to human reviewer answering the question correctly within 3 minutes, and (3) \emph{Wrong answer}: Rejected by human reviewers. Based on this feedback, annotators will refine their data until it passes the review process. To avoid wasting too much manual resources on low-quality data, we will terminate the review-revision cycle if the data has been revised more than five times without passing.

\xhdr{Mechanism Design}
To incentivize annotators to provide high-quality, challenging, and longer test data, our reward mechanism is set as follows. First, annotators can receive a base reward of \texttt{100} \texttt{CNY} only if the data passes the review process; no reward is given for data that does not pass. To encourage annotators to provide longer data, we offer additional length rewards of \texttt{20}, \texttt{40}, and \texttt{50} \texttt{CNY} for passed data in the length ranges $(32k, 64k]$, $(64k, 128k]$, and over $128k$, respectively (in word count). To motivate annotators to provide more difficult data, we define \emph{hard} set data as data where at least two out of three models do not answer correctly in automated review and the human reviewer is unable to solve it within 10 minutes; all other data is considered \emph{easy} data. For hard data, annotators can earn an additional difficulty reward of \texttt{50} \texttt{CNY}.
Each human expert is rewarded \texttt{25} \texttt{CNY} for reviewing each piece of data. We also conduct random checks on their reviews, and any human expert whose reviews repeatedly fail these checks will have all of their reviewing rewards revoked.

\subsection{Data Verification}
For a final check, we sample 70 test data and invite our authors to verify their correctness and whether they are Google-proofed~\cite{rein2023gpqa}.

\xhdr{Correctness} 
Check the selected answer based on the provided evidence to determine if it is correct, with all other options being incorrect. An answer is also deemed incorrect if there is any controversy, ambiguity, or reliance on subjective judgment.

\xhdr{Google-proof} Search for the answer to the question on the internet (Google). The data is considered Google-proof if the answer cannot be found within 15 minutes of searching.

Through our verification, we find that \texttt{68/70} of the data are completely correct, and \texttt{67/70} are Google-proofed. Therefore, we estimate that the error rate of our data is around 3\%, and the majority of the questions cannot be answered by memorizing existing data on the internet.
We review all the data to ensure that it does not contain any sensitive information related to privacy or copyrights.

\subsection{Data Statistics}
We categorize the 503 data entries in Longbench v2 based on their difficulty, length, and task types. According to the difficulty criteria defined in the previous section, 192 are classified as ``Easy'', while 311 are deemed ``Hard''. Based on word count, the data is divided into three groups: ``Short'' ($<$32k), ``Medium'' (32k-128k), and ``Long'' ($>$128k), containing 180, 215, and 108 entries, respectively, exhibiting a relatively balanced distribution.
For the data distribution across task types, please see Table~\ref{tb:stat}.
Also, the questions with answers A, B, C, and D account for approximately 19\%, 25\%, 30\%, and 26\% of the total, respectively, showing that the distribution of answers across the four options is relatively even.
We also analyze the proportion of data submissions rejected during manual review and find that 4\% of the submissions are rejected for \emph{illegal question}; 7\% are rejected for \emph{insufficient difficulty}; and 4\% are rejected for \emph{wrong answer}.

\section{Evaluation}

\subsection{Baselines}

\xhdr{Setup}
We evaluate 10 open-source LLMs, all of which have a context window size of 128,000 tokens, along with 7 proprietary LLMs.
We apply middle truncation as described in~\citet{bai2024longbench} for sequences exceeding the model's context window length.
Given the complex reasoning required by our test data, we adopt two evaluation settings: zero-shot and zero-shot + CoT. Following~\citet{rein2023gpqa}, in the CoT setting, the model is first prompted to generate a chain of thought~\cite{wei2022chain}, after which it is asked to produce the final answer based on the chain of thought.
For details on reproducing our results, please refer to Appendix~\ref{sec:setup}. 
For a fair comparison, the Qwen2.5 series models are evaluated without YaRN~\cite{peng2024yarn}. Their performance when combining YaRN are provided in Table~\ref{tb:exp_yarn}.
The code is available at \url{https://github.com/THUDM/LongBench}.

\begin{table*}[t]
\centering
\resizebox{\linewidth}{!}{
\begin{tabular}{p{5.7cm}|m{0.7cm}m{0.7cm}|m{0.7cm}m{0.7cm}|m{0.7cm}m{0.7cm}|m{0.7cm}m{0.7cm}|m{0.7cm}m{0.7cm}|m{0.7cm}m{0.7cm}}
\toprule
 &  & & \multicolumn{4}{|c}{\textbf{Difficulty}} & \multicolumn{6}{|c}{\textbf{Length (<32k; 32k-128k; >128k)$^\diamond$}} \\
\cmidrule(r){1-3} \cmidrule(lr){4-7} \cmidrule(l){8-13}
\textbf{Model} & \multicolumn{2}{c|}{\textbf{Overall}} & \multicolumn{2}{c|}{\textbf{Easy}} & \multicolumn{2}{c|}{\textbf{Hard}} & \multicolumn{2}{c|}{\textbf{Short}} & \multicolumn{2}{c|}{\textbf{Medium}} & \multicolumn{2}{c}{\textbf{Long}} \\ 
\midrule
\multicolumn{13}{l}{\emph{Open-source models}} \\
\texttt{GLM-4-9B-Chat} & 30.2 & \cellcolor{mygray}30.8 & 30.7 & \cellcolor{mygray}34.4 & 29.9 & \cellcolor{mygray}28.6 & 33.9 & \cellcolor{mygray}35.0 & 29.8 & \cellcolor{mygray}30.2 & 25.0 & \cellcolor{mygray}25.0 \\
\texttt{Llama-3.1-8B-Instruct} & 30.0 & \cellcolor{mygray}30.4 & 30.7 & \cellcolor{mygray}36.5 & 29.6 & \cellcolor{mygray}26.7 & 35.0 & \cellcolor{mygray}34.4 & 27.9 & \cellcolor{mygray}31.6 & 25.9 & \cellcolor{mygray}21.3 \\
\texttt{Llama-3.1-70B-Instruct} & 31.6 & \cellcolor{mygray}36.2 & 32.3 & \cellcolor{mygray}35.9 & 31.2 & \cellcolor{mygray}36.3 & 41.1 & \cellcolor{mygray}45.0 & 27.4 & \cellcolor{mygray}34.0 & 24.1 & \cellcolor{mygray}25.9 \\
\texttt{Llama-3.3-70B-Instruct} & 29.8 & \cellcolor{mygray}36.2 & 34.4 & \cellcolor{mygray}38.0 & 27.0 & \cellcolor{mygray}35.0 & 36.7 & \cellcolor{mygray}45.0 & 27.0 & \cellcolor{mygray}33.0 & 24.1 & \cellcolor{mygray}27.8 \\
\texttt{Llama-3.1-Nemotron-70B-Inst.} & 31.0 & \cellcolor{mygray}35.2 & 32.8 & \cellcolor{mygray}37.0 & 29.9 & \cellcolor{mygray}34.1 & 38.3 & \cellcolor{mygray}46.7 & 27.9 & \cellcolor{mygray}29.8 & 25.0 & \cellcolor{mygray}26.9 \\
\texttt{Qwen2.5-7B-Instruct} & 27.0 & \cellcolor{mygray}29.8 & 29.2 & \cellcolor{mygray}30.7 & 25.7 & \cellcolor{mygray}29.3 & 36.1 & \cellcolor{mygray}35.6 & 23.7 & \cellcolor{mygray}26.5 & 18.5 & \cellcolor{mygray}26.9 \\
\texttt{Qwen2.5-72B-Instruct} & \textbf{39.4} & \cellcolor{mygray}38.8 & \textbf{43.8} & \cellcolor{mygray}42.2 & \textbf{36.7} & \cellcolor{mygray}\textbf{36.7} & \textbf{44.4} & \cellcolor{mygray}\textbf{50.0} & \textbf{34.0} & \cellcolor{mygray}28.8 & \textbf{41.7} & \cellcolor{mygray}\textbf{39.8} \\
\texttt{Mistral-Large-Instruct-2407} & 26.6 & \cellcolor{mygray}33.6 & 29.7 & \cellcolor{mygray}34.4 & 24.8 & \cellcolor{mygray}33.1 & 37.8 & \cellcolor{mygray}41.1 & 19.5 & \cellcolor{mygray}31.2 & 22.2 & \cellcolor{mygray}25.9 \\
\texttt{Mistral-Large-Instruct-2411} & 34.4 & \cellcolor{mygray}\textbf{39.6} & 38.0 & \cellcolor{mygray}\textbf{43.8} & 32.2 & \cellcolor{mygray}37.0 & 41.7 & \cellcolor{mygray}46.1 & 30.7 & \cellcolor{mygray}\textbf{34.9} & 29.6 & \cellcolor{mygray}38.0 \\
\texttt{c4ai-command-r-plus-08-2024} & 27.8 & \cellcolor{mygray}31.6 & 30.2 & \cellcolor{mygray}34.4 & 26.4 & \cellcolor{mygray}29.9 & 36.7 & \cellcolor{mygray}39.4 & 23.7 & \cellcolor{mygray}24.2 & 21.3 & \cellcolor{mygray}33.3 \\ 
\midrule
\multicolumn{13}{l}{\emph{Proprietary models}} \\
\texttt{GLM-4-Plus} & 44.3 & \cellcolor{mygray}46.1 & 47.4 & \cellcolor{mygray}52.1 & 42.4 & \cellcolor{mygray}42.4 & 50.0 & \cellcolor{mygray}53.3 & 46.5 & \cellcolor{mygray}44.7 & 30.6 & \cellcolor{mygray}37.0 \\
\texttt{GPT-4o-mini-2024-07-18} & 29.3 & \cellcolor{mygray}32.4 & 31.1 & \cellcolor{mygray}32.6 & 28.2 & \cellcolor{mygray}32.2 & 31.8 & \cellcolor{mygray}34.8 & 28.6 & \cellcolor{mygray}31.6 & 26.2 & \cellcolor{mygray}29.9 \\
\texttt{GPT-4o-2024-08-06} & 50.1 & \cellcolor{mygray}51.2 & 57.4 & \cellcolor{mygray}57.9 & 45.6 & \cellcolor{mygray}47.1 & 53.3 & \cellcolor{mygray}53.9 & 52.4 & \cellcolor{mygray}\textbf{50.7} & 40.2 & \cellcolor{mygray}47.7 \\
\texttt{GPT-4o-2024-11-20} & 46.0 & \cellcolor{mygray}51.4 & 50.8 & \cellcolor{mygray}54.2 & 43.0 & \cellcolor{mygray}49.7 & 47.5 & \cellcolor{mygray}59.6 & 47.9 & \cellcolor{mygray}48.6 & 39.8 & \cellcolor{mygray}43.5 \\
\texttt{o1-mini-2024-09-12} & 37.8 & \cellcolor{mygray}38.9 & 38.9 & \cellcolor{mygray}42.6 & 37.1 & \cellcolor{mygray}36.6 & 48.6 & \cellcolor{mygray}48.9 & 33.3 & \cellcolor{mygray}32.9 & 28.6 & \cellcolor{mygray}34.3 \\
\texttt{o1-preview-2024-09-12} & \textbf{57.7} & \cellcolor{mygray}\textbf{56.2} & \textbf{66.8} & \cellcolor{mygray}\textbf{58.9} & \textbf{52.1} & \cellcolor{mygray}\textbf{54.6} & \textbf{62.6} & \cellcolor{mygray}\textbf{64.6} & \textbf{53.5} & \cellcolor{mygray}50.2 & \textbf{58.1} & \cellcolor{mygray}\textbf{54.3} \\
\texttt{Claude-3.5-Sonnet-20241022} & 41.0 & \cellcolor{mygray}46.7 & 46.9 & \cellcolor{mygray}55.2 & 37.3 & \cellcolor{mygray}41.5 & 46.1 & \cellcolor{mygray}53.9 & 38.6 & \cellcolor{mygray}41.9 & 37.0 & \cellcolor{mygray}44.4\\
\midrule
\cellcolor{mypink}\emph{Human$^*$} & \multicolumn{2}{c|}{\cellcolor{mypink}53.7} & \multicolumn{2}{c|}{\cellcolor{mypink}100} & \multicolumn{2}{c|}{\cellcolor{mypink}25.1} & \multicolumn{2}{c|}{\cellcolor{mypink}47.2} & \multicolumn{2}{c|}{\cellcolor{mypink}59.1} & \multicolumn{2}{c}{\cellcolor{mypink}53.7} \\
\bottomrule
\end{tabular}
}
\caption{Evaluation results (\%) on LongBench v2. Results under \colorbox{mygray}{CoT} prompting are highlighted with a gray background. Note that random guessing yields a baseline score of 25\%. To account for model responses and human responses that do not yield a valid choice, we report the \emph{compensated} results in Table~\ref{tb:exp_comp}, where these cases are counted towards the accuracy with a random probability of 25\%. $^*$: The human expert's accuracy is based on their performance within a 15-minute time limit, after which they are allowed to respond with ``I don't know the answer''. This occurred for 8\% of the total test data. $^\diamond$: Models do not show lower scores on subsets with longer length ranges because the distribution of tasks differs significantly across each length range (Figure~\ref{fig:length}).}
\label{tb:exp}
\end{table*}

\xhdr{Results}
We report the evaluation results along with human expert performance in Table~\ref{tb:exp}. The results under the CoT evaluation setting are highlighted with a gray background, while the highest scores among open-source models and proprietary models are in bold.
The results indicate that LongBench v2 presents a significant challenge to the current model---The best-performing o1-preview model achieves only 57.7\% accuracy, which is 4\% higher than the performance of human experts under a 15-minute time limit. Additionally, the scaling law effect on our benchmark is striking: smaller models such as GLM-4-9B-Chat, Qwen2.5-7B-Instruct, and GPT-4o-mini perform poorly in our tests that require deep understanding and reasoning over long contexts, with accuracy around 30\%. In contrast, their larger counterparts like GLM-4-Plus, Qwen2.5-72B-Instruct, and GPT-4o show a notable improvement, achieving overall accuracy around or above 40\%.
Similar to reasoning tasks in mathematics and coding~\cite{wei2022chain,sprague2024cot,o1-preview}, we also find that incorporating explicit reasoning in the model’s responses significantly improves its performance in our long-context reasoning tests.
This includes the use of CoT, which results in an average 3.4\% improvement for open-source models. Additionally, scaling test-time compute with longer reasoning thought shows further improvements, with o1-preview vs. GPT-4o (+7.6\%) and o1-mini vs. GPT-4o-mini (+8.5\%).
From the performance across different length intervals, compared to human, the models perform best on data $<$32k (Short), with the best-performing model surpassing human performance by 15.4\%. However, even the top model shows a 5.6\% performance gap compared to human accuracy in the 32k-128k data length range. This highlights the importance of developing methods to maintain strong reasoning capabilities under longer contexts.

\begin{figure}[t]
    \centering
    \includegraphics[width=0.9\linewidth]{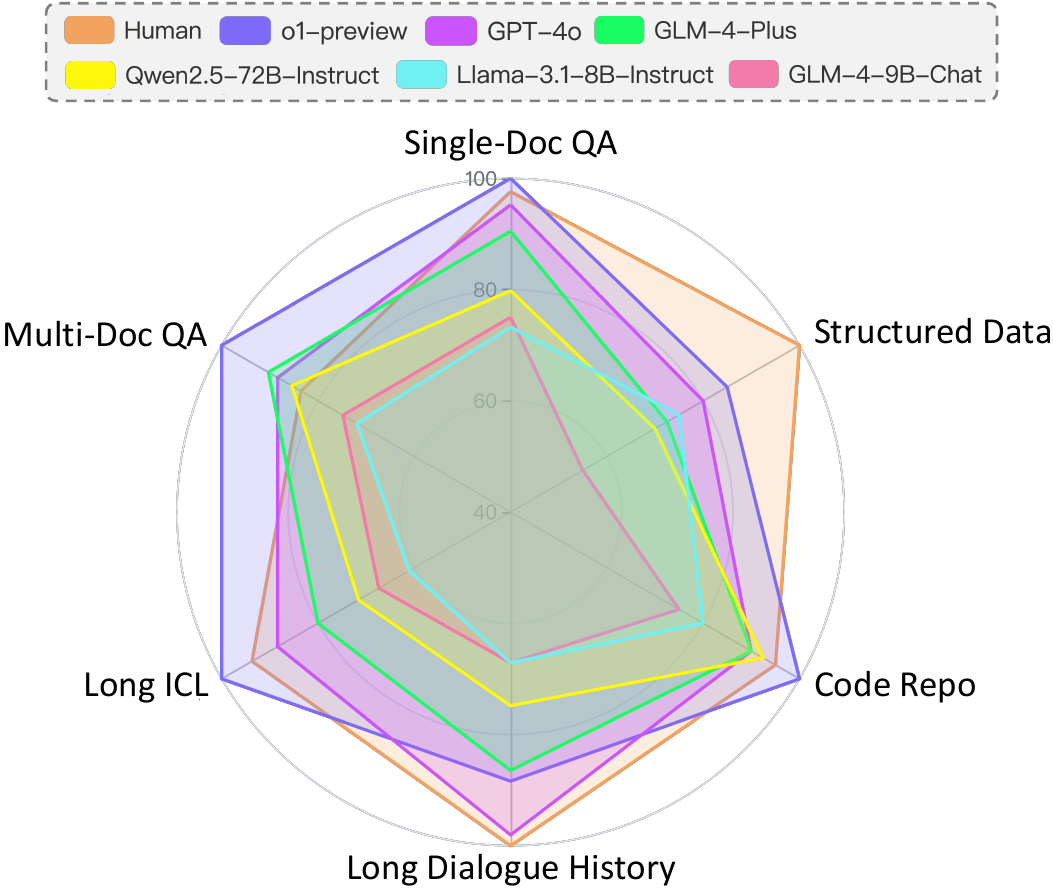}
    \caption{Average scores across tasks, normalized by the highest score on each task. All scores are evaluated in the zero-shot + CoT setting, except for o1-preview, since it latently performs CoT under zero-shot prompting.}
    \label{fig:radar}
\end{figure}

To better distinguish the capability of the models across tasks, we present the performance charts of several representative models across tasks in Figure~\ref{fig:radar}.
We find that the performance gap between LLMs and humans is largest on long structured data understanding tasks, whereas, on single-doc and multi-doc QA tasks, the models perform at par with or even surpass human levels.
We hypothesize that this is because the models have seen much more document-type data compared to long structured data during long context training, resulting in poorer understanding of the latter.
Compared to GPT-4o, we observe that through integrating more thinking steps during inference, o1-preview shows superior performance on multi-doc QA, long in-context learning, and code repository understanding tasks, with a substantial lead over other models.

\subsection{Retrieval-Augmented Baselines}

\begin{figure}[t]
    \centering
    \includegraphics[width=\linewidth]{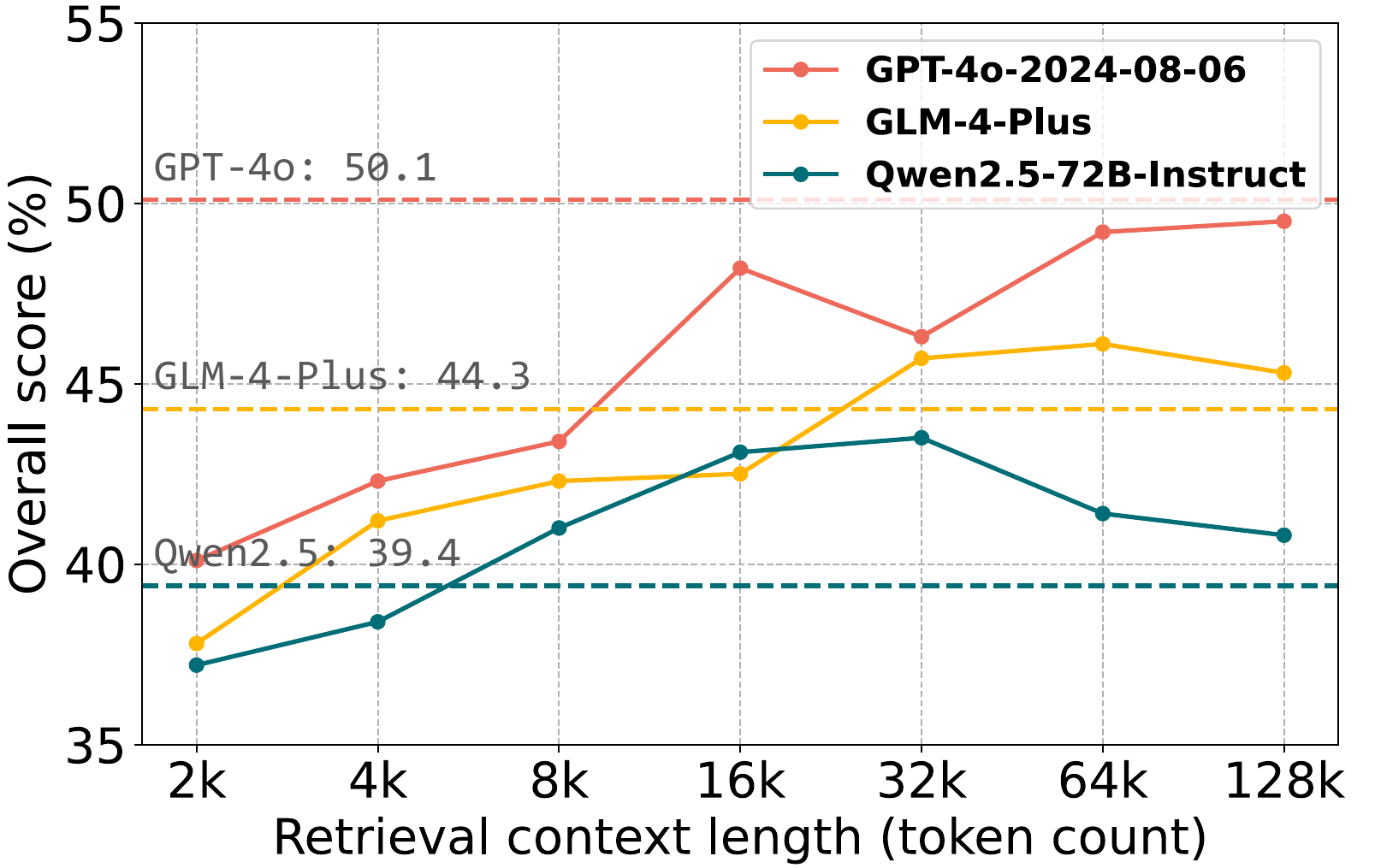}
    \caption{RAG performance across different context lengths, varied by including the top 4, 8, 16, 32, 64, 128, and 256 chunks of 512 tokens. The horizontal line show the overall score of each model without RAG at a full context length of 128k tokens.}
    \label{fig:rag}
\end{figure}

Based on recent studies~\cite{jiang2024longrag,jin2024long,leng2024long}, we explore incorporating retrieval-augmented generation (RAG,~\citet{lewis2020retrieval}) into long-context LLM and evaluate its performance on LongBench v2.
We first split the long context into chunks of 512 tokens with GLM-4-9B tokenizer. Then, we use \href{https://open.bigmodel.cn/pricing}{Zhipu Embedding-3} to encode the query, i.e., the concatenation of the question and choices, and the chunks, and sort the chunks based on embedding similarity.
During evaluation, we retrieve the top-$N$ most similar chunks and concatenate them in their original order to form the context input for the model. The model is then prompted to answer the question in a zero-shot setting. For each evaluated model, we take $N = 4, 8, 16, 32, 64, 128, \text{and}\ 256$, and the evaluation results form a curve presented in Figure~\ref{fig:rag}.

We observe that Qwen2.5 and GLM-4-Plus show no significant improvement as the retrieval context length increases beyond 32k. Both models perform better at a 32k retrieval context length compared to using the entire 128k context window without RAG, with Qwen2.5 showing a notable improvement of +4.1\%. 
In contrast, only GPT-4o effectively leverages longer retrieval context lengths, achieving the best RAG performance at 128k, while still lagging behind its overall score without RAG (-0.6\%).
These findings suggest that Qwen2.5 and GLM-4-Plus fall short in effectively utilizing and reasoning with information in context windows longer than 32k compared to GPT-4o.
In addition, these experiments also confirm that the questions in LongBench v2 are challenging and cannot be solved solely through retrieval.

\subsection{Measuring Memorization of Context}

\begin{table}[t]
\centering
\resizebox{\linewidth}{!}{
\begin{tabular}{lccccccc}
\toprule
\textbf{Model} & \textbf{Avg} & \textbf{I} & \textbf{II} & \textbf{III} & \textbf{IV} & \textbf{V} & \textbf{VI} \\
\midrule
\cellcolor{mygray}\texttt{GLM-4-9B-Chat} & \cellcolor{mygray}30.2 & \cellcolor{mygray}30.9 & \cellcolor{mygray}27.2 & \cellcolor{mygray}33.3 & \cellcolor{mygray}38.5 & \cellcolor{mygray}28.0 & \cellcolor{mygray}24.2 \\
\quad w/o context & 26.2 & 30.9 & 21.6 & 18.5 & 30.8 & 34.0 & 21.2 \\
\midrule
\cellcolor{mygray}\texttt{Llama-3.1-8B-Inst.} & \cellcolor{mygray}30.0 & \cellcolor{mygray}34.9 & \cellcolor{mygray}30.4 & \cellcolor{mygray}23.5 & \cellcolor{mygray}17.9 & \cellcolor{mygray}32.0 & \cellcolor{mygray}30.3 \\
\quad w/o context & 25.8 & 31.4 & 26.4 & 24.7 & 23.1 & 22.0 & 6.1 \\
\midrule
\cellcolor{mygray}\texttt{Qwen2.5-72B-Inst.} & \cellcolor{mygray}39.4 & \cellcolor{mygray}40.6 & \cellcolor{mygray}35.2 & \cellcolor{mygray}42.0 & \cellcolor{mygray}25.6 & \cellcolor{mygray}50.0 & \cellcolor{mygray}42.4 \\
\quad w/o context & 30.0 & 33.7 & 31.2 & 25.9 & 28.2 & 34.0 & 12.1 \\
\midrule
\cellcolor{mygray}\texttt{GLM-4-Plus} & \cellcolor{mygray}44.3 & \cellcolor{mygray}41.7 & \cellcolor{mygray}42.4 & \cellcolor{mygray}46.9 & \cellcolor{mygray}51.3 & \cellcolor{mygray}46.0 & \cellcolor{mygray}48.5 \\
\quad w/o context & 27.6 & 33.7 & 27.2 & 25.9 & 10.3 & 38.0 & 6.1 \\
\midrule
\cellcolor{mygray}\texttt{GPT-4o} & \cellcolor{mygray}50.1 & \cellcolor{mygray}48.6 & \cellcolor{mygray}44.0 & \cellcolor{mygray}58.0 & \cellcolor{mygray}46.2 & \cellcolor{mygray}56.0 & \cellcolor{mygray}51.5 \\
\quad w/o context & 33.1 & 40.0 & 25.6 & 32.1 & 38.5 & 34.0 & 18.2 \\
\bottomrule
\end{tabular}
}
\caption{Scores (\%) across 6 tasks: \emph{I. Single-Doc QA}, \emph{II. Multi-Doc QA}, \emph{III. Long ICL}, \emph{IV. Dialogue History}, \emph{V. Code Repo}, and \emph{VI. Structured Data}.}
\label{tb:mem}
\end{table}

For an effective long-context benchmark, it is essential to ensure that LLMs cannot rely solely on memorizing previously seen data to answer questions. 
This necessitates the models to actively read and comprehend the provided long material in order to solve the problems.
Following~\citet{bai2024longbench}, we also evaluate the models' performance when providing only the questions, without the accompanying long context.
The performance comparison between with (w/) and without (w/o) the context is presented in Table~\ref{tb:mem}.
As shown, without context, most models achieve an overall accuracy ranging from 25\% to 30\%, which is comparable to random guessing. When comparing scores across different tasks, the memorization effect appears minimal for tasks II, III, and VI.
The models perform best without context on tasks I and V, likely because they may have seen some of the documents, novels, or code repositories during training.

\section{Conclusion}
Our work introduces LongBench v2, a challenging multitask benchmark for long-context understanding and reasoning, carefully annotated and reviewed by human experts.
LongBench v2 presents an equal challenge to both humans and state-of-the-art AI systems, with human performance at 50.1\% and the best LLM achieving 57.7\% accuracy, providing a reliable evaluation standard for the development of future superhuman AI systems.
Our evaluation results also bring forward insights into the impact of scaling inference-time compute and RAG in long-context reasoning.

\section{Limitations}
We acknowledge certain limitations in our work, which we outline below:
1. \textbf{Benchmark size}: The benchmark's size may not be sufficiently large. While this can be seen as an advantage for quick evaluation, it could also lead to less stable results that are more vulnerable to randomness. Due to resource constraints, we are unable to expand the dataset at this time. Collecting the current 503 high-quality samples cost us 100,000 CNY and took more than two months.
2. \textbf{Language}: The current dataset is limited to English only. As a result, our benchmark does not yet capture the performance of models across multiple languages.
3. \textbf{Length distribution inconsistencies}: The length distribution across different tasks is uneven, with certain tasks concentrated around specific lengths. These differences in task distributions across length ranges make it difficult to provide a fair comparison of a single model's performance across length intervals. We recommend conducting comparisons between models on a per-interval basis. For instance, model A may outperform Model B in the short length range, while model B may outperform model A in the long length range. This would suggest that model B is better at handling longer tasks than model A.

\section*{Acknowledgements}
We would like to express our gratitude to our annotation workers for their dedicated contributions. The authors also extend their thanks to Zijun Yao for his assistance in maintaining the platform, and to Yuze He for his valuable suggestions on the paper.

\bibliography{custom}

\begin{thebibliography}{73}
\providecommand{\natexlab}[1]{#1}

\bibitem[{Agarwal et~al.(2024)Agarwal, Singh, Zhang, Bohnet, Rosias, Chan, Zhang, Anand, Abbas, Nova et~al.}]{agarwal2024many}
Rishabh Agarwal, Avi Singh, Lei~M Zhang, Bernd Bohnet, Luis Rosias, Stephanie Chan, Biao Zhang, Ankesh Anand, Zaheer Abbas, Azade Nova, et~al. 2024.
\newblock Many-shot in-context learning.
\newblock \emph{arXiv preprint arXiv:2404.11018}.

\bibitem[{An et~al.(2024)An, Gong, Zhong, Zhao, Li, Zhang, Kong, and Qiu}]{an2024leval}
Chenxin An, Shansan Gong, Ming Zhong, Xingjian Zhao, Mukai Li, Jun Zhang, Lingpeng Kong, and Xipeng Qiu. 2024.
\newblock \href {https://doi.org/10.18653/v1/2024.acl-long.776} {{L}-eval: Instituting standardized evaluation for long context language models}.
\newblock In \emph{Proceedings of the 62nd Annual Meeting of the Association for Computational Linguistics (Volume 1: Long Papers)}, pages 14388--14411, Bangkok, Thailand. Association for Computational Linguistics.

\bibitem[{Anthropic(2024)}]{claude-3-5}
Anthropic. 2024.
\newblock \href {https://www.anthropic.com/news/claude-3-5-sonnet} {Anthropic: Introducing claude 3.5 sonnet}.

\bibitem[{Bai et~al.(2023)Bai, Lv, Li, and Hou}]{bai2023answering}
Yushi Bai, Xin Lv, Juanzi Li, and Lei Hou. 2023.
\newblock Answering complex logical queries on knowledge graphs via query computation tree optimization.
\newblock In \emph{International Conference on Machine Learning}, pages 1472--1491. PMLR.

\bibitem[{Bai et~al.(2024{\natexlab{a}})Bai, Lv, Zhang, He, Qi, Hou, Tang, Dong, and Li}]{bai2024longalign}
Yushi Bai, Xin Lv, Jiajie Zhang, Yuze He, Ji~Qi, Lei Hou, Jie Tang, Yuxiao Dong, and Juanzi Li. 2024{\natexlab{a}}.
\newblock \href {https://doi.org/10.18653/v1/2024.findings-emnlp.74} {{L}ong{A}lign: A recipe for long context alignment of large language models}.
\newblock In \emph{Findings of the Association for Computational Linguistics: EMNLP 2024}, pages 1376--1395, Miami, Florida, USA. Association for Computational Linguistics.

\bibitem[{Bai et~al.(2024{\natexlab{b}})Bai, Lv, Zhang, Lyu, Tang, Huang, Du, Liu, Zeng, Hou, Dong, Tang, and Li}]{bai2024longbench}
Yushi Bai, Xin Lv, Jiajie Zhang, Hongchang Lyu, Jiankai Tang, Zhidian Huang, Zhengxiao Du, Xiao Liu, Aohan Zeng, Lei Hou, Yuxiao Dong, Jie Tang, and Juanzi Li. 2024{\natexlab{b}}.
\newblock \href {https://doi.org/10.18653/v1/2024.acl-long.172} {{L}ong{B}ench: A bilingual, multitask benchmark for long context understanding}.
\newblock In \emph{Proceedings of the 62nd Annual Meeting of the Association for Computational Linguistics (Volume 1: Long Papers)}, pages 3119--3137, Bangkok, Thailand. Association for Computational Linguistics.

\bibitem[{Bai et~al.(2024{\natexlab{c}})Bai, Ying, Cao, Lv, He, Wang, Yu, Zeng, Xiao, Lyu et~al.}]{bai2024benchmarking}
Yushi Bai, Jiahao Ying, Yixin Cao, Xin Lv, Yuze He, Xiaozhi Wang, Jifan Yu, Kaisheng Zeng, Yijia Xiao, Haozhe Lyu, et~al. 2024{\natexlab{c}}.
\newblock Benchmarking foundation models with language-model-as-an-examiner.
\newblock \emph{Advances in Neural Information Processing Systems}, 36.

\bibitem[{Bai et~al.(2024{\natexlab{d}})Bai, Zhang, Lv, Zheng, Zhu, Hou, Dong, Tang, and Li}]{bai2024longwriter}
Yushi Bai, Jiajie Zhang, Xin Lv, Linzhi Zheng, Siqi Zhu, Lei Hou, Yuxiao Dong, Jie Tang, and Juanzi Li. 2024{\natexlab{d}}.
\newblock Longwriter: Unleashing 10,000+ word generation from long context llms.
\newblock \emph{arXiv preprint arXiv:2408.07055}.

\bibitem[{Bogomolov et~al.(2024)Bogomolov, Eliseeva, Galimzyanov, Glukhov, Shapkin, Tigina, Golubev, Kovrigin, van Deursen, Izadi et~al.}]{bogomolov2024long}
Egor Bogomolov, Aleksandra Eliseeva, Timur Galimzyanov, Evgeniy Glukhov, Anton Shapkin, Maria Tigina, Yaroslav Golubev, Alexander Kovrigin, Arie van Deursen, Maliheh Izadi, et~al. 2024.
\newblock Long code arena: a set of benchmarks for long-context code models.
\newblock \emph{arXiv preprint arXiv:2406.11612}.

\bibitem[{Cao et~al.(2022)Cao, Shi, Pan, Nie, Xiang, Hou, Li, He, and Zhang}]{cao2022kqa}
Shulin Cao, Jiaxin Shi, Liangming Pan, Lunyiu Nie, Yutong Xiang, Lei Hou, Juanzi Li, Bin He, and Hanwang Zhang. 2022.
\newblock Kqa pro: A dataset with explicit compositional programs for complex question answering over knowledge base.
\newblock In \emph{Proceedings of the 60th Annual Meeting of the Association for Computational Linguistics (Volume 1: Long Papers)}, pages 6101--6119.

\bibitem[{{Cohere For AI}(2024)}]{cohere_for_ai_2024}
{Cohere For AI}. 2024.
\newblock \href {https://doi.org/10.57967/hf/3135} {c4ai-command-r-plus-08-2024}.

\bibitem[{Dasigi et~al.(2021)Dasigi, Lo, Beltagy, Cohan, Smith, and Gardner}]{dasigi2021dataset}
Pradeep Dasigi, Kyle Lo, Iz~Beltagy, Arman Cohan, Noah~A Smith, and Matt Gardner. 2021.
\newblock A dataset of information-seeking questions and answers anchored in research papers.
\newblock In \emph{Proceedings of the 2021 Conference of the North American Chapter of the Association for Computational Linguistics: Human Language Technologies}, pages 4599--4610.

\bibitem[{Demszky et~al.(2020)Demszky, Movshovitz-Attias, Ko, Cowen, Nemade, and Ravi}]{demszky2020goemotions}
Dorottya Demszky, Dana Movshovitz-Attias, Jeongwoo Ko, Alan Cowen, Gaurav Nemade, and Sujith Ravi. 2020.
\newblock Goemotions: A dataset of fine-grained emotions.
\newblock In \emph{Proceedings of the 58th Annual Meeting of the Association for Computational Linguistics}, pages 4040--4054.

\bibitem[{Ding et~al.(2021)Ding, Xu, Chen, Wang, Han, Xie, Zheng, and Liu}]{ding2021few}
Ning Ding, Guangwei Xu, Yulin Chen, Xiaobin Wang, Xu~Han, Pengjun Xie, Haitao Zheng, and Zhiyuan Liu. 2021.
\newblock Few-nerd: A few-shot named entity recognition dataset.
\newblock In \emph{Proceedings of the 59th Annual Meeting of the Association for Computational Linguistics and the 11th International Joint Conference on Natural Language Processing (Volume 1: Long Papers)}, pages 3198--3213.

\bibitem[{Dong et~al.(2024)Dong, Tang, Li, Zhao, and Wen}]{dong2024bamboo}
Zican Dong, Tianyi Tang, Junyi Li, Wayne~Xin Zhao, and Ji-Rong Wen. 2024.
\newblock Bamboo: A comprehensive benchmark for evaluating long text modeling capacities of large language models.
\newblock In \emph{Proceedings of the 2024 Joint International Conference on Computational Linguistics, Language Resources and Evaluation (LREC-COLING 2024)}, pages 2086--2099.

\bibitem[{Dua et~al.(2019)Dua, Wang, Dasigi, Stanovsky, Singh, and Gardner}]{dua2019drop}
Dheeru Dua, Yizhong Wang, Pradeep Dasigi, Gabriel Stanovsky, Sameer Singh, and Matt Gardner. 2019.
\newblock Drop: A reading comprehension benchmark requiring discrete reasoning over paragraphs.
\newblock In \emph{Proceedings of the 2019 Conference of the North American Chapter of the Association for Computational Linguistics: Human Language Technologies, Volume 1 (Long and Short Papers)}, pages 2368--2378.

\bibitem[{Dubey et~al.(2024)Dubey, Jauhri, Pandey, Kadian, Al-Dahle, Letman, Mathur, Schelten, Yang, Fan et~al.}]{dubey2024llama}
Abhimanyu Dubey, Abhinav Jauhri, Abhinav Pandey, Abhishek Kadian, Ahmad Al-Dahle, Aiesha Letman, Akhil Mathur, Alan Schelten, Amy Yang, Angela Fan, et~al. 2024.
\newblock The llama 3 herd of models.
\newblock \emph{arXiv preprint arXiv:2407.21783}.

\bibitem[{Fu et~al.(2024)Fu, Panda, Niu, Yue, Hajishirzi, Kim, and Peng}]{pmlr-v235-fu24d}
Yao Fu, Rameswar Panda, Xinyao Niu, Xiang Yue, Hannaneh Hajishirzi, Yoon Kim, and Hao Peng. 2024.
\newblock Data engineering for scaling language models to 128{K} context.
\newblock In \emph{Proceedings of the 41st International Conference on Machine Learning}, volume 235 of \emph{Proceedings of Machine Learning Research}, pages 14125--14134. PMLR.

\bibitem[{Gao et~al.(2024)Gao, Wettig, Yen, and Chen}]{gao2024train}
Tianyu Gao, Alexander Wettig, Howard Yen, and Danqi Chen. 2024.
\newblock How to train long-context language models (effectively).
\newblock \emph{arXiv preprint arXiv:2410.02660}.

\bibitem[{GLM et~al.(2024)GLM, Zeng, Xu, Wang, Zhang, Yin, Rojas, Feng, Zhao, Lai, Yu, Wang, Sun, Zhang, Cheng, Gui, Tang, Zhang, Li, Zhao, Wu, Zhong, Liu, Huang, Zhang, Zheng, Lu, Duan, Zhang, Cao, Yang, Tam, Zhao, Liu, Xia, Zhang, Gu, Lv, Liu, Liu, Yang, Song, Zhang, An, Xu, Niu, Yang, Li, Bai, Dong, Qi, Wang, Yang, Du, Hou, and Wang}]{glm2024chatglm}
Team GLM, Aohan Zeng, Bin Xu, Bowen Wang, Chenhui Zhang, Da~Yin, Diego Rojas, Guanyu Feng, Hanlin Zhao, Hanyu Lai, Hao Yu, Hongning Wang, Jiadai Sun, Jiajie Zhang, Jiale Cheng, Jiayi Gui, Jie Tang, Jing Zhang, Juanzi Li, Lei Zhao, Lindong Wu, Lucen Zhong, Mingdao Liu, Minlie Huang, Peng Zhang, Qinkai Zheng, Rui Lu, Shuaiqi Duan, Shudan Zhang, Shulin Cao, Shuxun Yang, Weng~Lam Tam, Wenyi Zhao, Xiao Liu, Xiao Xia, Xiaohan Zhang, Xiaotao Gu, Xin Lv, Xinghan Liu, Xinyi Liu, Xinyue Yang, Xixuan Song, Xunkai Zhang, Yifan An, Yifan Xu, Yilin Niu, Yuantao Yang, Yueyan Li, Yushi Bai, Yuxiao Dong, Zehan Qi, Zhaoyu Wang, Zhen Yang, Zhengxiao Du, Zhenyu Hou, and Zihan Wang. 2024.
\newblock Chatglm: A family of large language models from glm-130b to glm-4 all tools.
\newblock \emph{arXiv preprint arXiv:2406.12793}.

\bibitem[{Hsieh et~al.(2024)Hsieh, Sun, Kriman, Acharya, Rekesh, Jia, Zhang, and Ginsburg}]{hsieh2024ruler}
Cheng-Ping Hsieh, Simeng Sun, Samuel Kriman, Shantanu Acharya, Dima Rekesh, Fei Jia, Yang Zhang, and Boris Ginsburg. 2024.
\newblock Ruler: What's the real context size of your long-context language models?
\newblock \emph{arXiv preprint arXiv:2404.06654}.

\bibitem[{Huang et~al.(2024)Huang, Li, Lam, Liang, Wang, Yuan, Jiao, Wang, Tu, and Lyu}]{huang2024far}
Jen-tse Huang, Eric~John Li, Man~Ho Lam, Tian Liang, Wenxuan Wang, Youliang Yuan, Wenxiang Jiao, Xing Wang, Zhaopeng Tu, and Michael~R Lyu. 2024.
\newblock How far are we on the decision-making of llms? evaluating llms' gaming ability in multi-agent environments.
\newblock \emph{arXiv preprint arXiv:2403.11807}.

\bibitem[{Huang et~al.(2021)Huang, Cao, Parulian, Ji, and Wang}]{huang2021efficient}
Luyang Huang, Shuyang Cao, Nikolaus Parulian, Heng Ji, and Lu~Wang. 2021.
\newblock Efficient attentions for long document summarization.
\newblock In \emph{Proceedings of the 2021 Conference of the North American Chapter of the Association for Computational Linguistics: Human Language Technologies}, pages 1419--1436.

\bibitem[{Jiang et~al.(2023)Jiang, Sablayrolles, Mensch, Bamford, Chaplot, Casas, Bressand, Lengyel, Lample, Saulnier et~al.}]{jiang2023mistral}
Albert~Q Jiang, Alexandre Sablayrolles, Arthur Mensch, Chris Bamford, Devendra~Singh Chaplot, Diego de~las Casas, Florian Bressand, Gianna Lengyel, Guillaume Lample, Lucile Saulnier, et~al. 2023.
\newblock Mistral 7b.
\newblock \emph{arXiv preprint arXiv:2310.06825}.

\bibitem[{Jiang et~al.(2024)Jiang, Ma, and Chen}]{jiang2024longrag}
Ziyan Jiang, Xueguang Ma, and Wenhu Chen. 2024.
\newblock Longrag: Enhancing retrieval-augmented generation with long-context llms.
\newblock \emph{arXiv preprint arXiv:2406.15319}.

\bibitem[{Jin et~al.(2024)Jin, Yoon, Han, and Arik}]{jin2024long}
Bowen Jin, Jinsung Yoon, Jiawei Han, and Sercan~O Arik. 2024.
\newblock Long-context llms meet rag: Overcoming challenges for long inputs in rag.
\newblock \emph{arXiv preprint arXiv:2410.05983}.

\bibitem[{Kamradt(2023)}]{needleinhaystack}
Greg Kamradt. 2023.
\newblock \href {https://github.com/gkamradt/LLMTest_NeedleInAHaystack} {Needle in a haystack - pressure testing llms}.
\newblock \url{https://github.com/gkamradt/LLMTest_NeedleInAHaystack}.

\bibitem[{Ko{\v{c}}isk{\`y} et~al.(2018)Ko{\v{c}}isk{\`y}, Schwarz, Blunsom, Dyer, Hermann, Melis, and Grefenstette}]{kovcisky2018narrativeqa}
Tom{\'a}{\v{s}} Ko{\v{c}}isk{\`y}, Jonathan Schwarz, Phil Blunsom, Chris Dyer, Karl~Moritz Hermann, G{\'a}bor Melis, and Edward Grefenstette. 2018.
\newblock The narrativeqa reading comprehension challenge.
\newblock \emph{Transactions of the Association for Computational Linguistics}, 6:317--328.

\bibitem[{Krishna et~al.(2024)Krishna, Krishna, Mohananey, Schwarcz, Stambler, Upadhyay, and Faruqui}]{krishna2024fact}
Satyapriya Krishna, Kalpesh Krishna, Anhad Mohananey, Steven Schwarcz, Adam Stambler, Shyam Upadhyay, and Manaal Faruqui. 2024.
\newblock Fact, fetch, and reason: A unified evaluation of retrieval-augmented generation.
\newblock \emph{arXiv preprint arXiv:2409.12941}.

\bibitem[{Kuratov et~al.(2024)Kuratov, Bulatov, Anokhin, Rodkin, Sorokin, Sorokin, and Burtsev}]{kuratov2024babilong}
Yuri Kuratov, Aydar Bulatov, Petr Anokhin, Ivan Rodkin, Dmitry Sorokin, Artyom Sorokin, and Mikhail Burtsev. 2024.
\newblock Babilong: Testing the limits of llms with long context reasoning-in-a-haystack.
\newblock \emph{arXiv preprint arXiv:2406.10149}.

\bibitem[{Laban et~al.(2024)Laban, Fabbri, Xiong, and Wu}]{laban2024summary}
Philippe Laban, Alexander~Richard Fabbri, Caiming Xiong, and Chien-Sheng Wu. 2024.
\newblock Summary of a haystack: A challenge to long-context llms and rag systems.
\newblock In \emph{Proceedings of the 2024 Conference on Empirical Methods in Natural Language Processing}, pages 9885--9903.

\bibitem[{Leng et~al.(2024)Leng, Portes, Havens, Zaharia, and Carbin}]{leng2024long}
Quinn Leng, Jacob Portes, Sam Havens, Matei Zaharia, and Michael Carbin. 2024.
\newblock Long context rag performance of large language models.
\newblock \emph{arXiv preprint arXiv:2411.03538}.

\bibitem[{Lewis et~al.(2020)Lewis, Perez, Piktus, Petroni, Karpukhin, Goyal, K{\"u}ttler, Lewis, Yih, Rockt{\"a}schel et~al.}]{lewis2020retrieval}
Patrick Lewis, Ethan Perez, Aleksandra Piktus, Fabio Petroni, Vladimir Karpukhin, Naman Goyal, Heinrich K{\"u}ttler, Mike Lewis, Wen-tau Yih, Tim Rockt{\"a}schel, et~al. 2020.
\newblock Retrieval-augmented generation for knowledge-intensive nlp tasks.
\newblock \emph{Advances in Neural Information Processing Systems}, 33:9459--9474.

\bibitem[{Li et~al.(2024)Li, Jiang, Huang, Beigi, Zhao, Tan, Bhattacharjee, Jiang, Chen, Wu et~al.}]{li2024generation}
Dawei Li, Bohan Jiang, Liangjie Huang, Alimohammad Beigi, Chengshuai Zhao, Zhen Tan, Amrita Bhattacharjee, Yuxuan Jiang, Canyu Chen, Tianhao Wu, et~al. 2024.
\newblock From generation to judgment: Opportunities and challenges of llm-as-a-judge.
\newblock \emph{arXiv preprint arXiv:2411.16594}.

\bibitem[{Li et~al.(2023)Li, Wang, Zheng, and Zhang}]{li2023loogle}
Jiaqi Li, Mengmeng Wang, Zilong Zheng, and Muhan Zhang. 2023.
\newblock Loogle: Can long-context language models understand long contexts?
\newblock \emph{arXiv preprint arXiv:2311.04939}.

\bibitem[{Liu et~al.(2023)Liu, Xu, and McAuley}]{liu2023repobench}
Tianyang Liu, Canwen Xu, and Julian McAuley. 2023.
\newblock Repobench: Benchmarking repository-level code auto-completion systems.
\newblock \emph{arXiv preprint arXiv:2306.03091}.

\bibitem[{Liu et~al.(2024)Liu, Dong, Hu, and Chu}]{liu2024longgenbench}
Xiang Liu, Peijie Dong, Xuming Hu, and Xiaowen Chu. 2024.
\newblock Longgenbench: Long-context generation benchmark.
\newblock In \emph{Findings of the Association for Computational Linguistics: EMNLP 2024}, pages 865--883.

\bibitem[{Novikova et~al.(2017)Novikova, Du{\v{s}}ek, Curry, and Rieser}]{novikova2017we}
Jekaterina Novikova, Ond{\v{r}}ej Du{\v{s}}ek, Amanda~Cercas Curry, and Verena Rieser. 2017.
\newblock Why we need new evaluation metrics for nlg.
\newblock In \emph{Proceedings of the 2017 Conference on Empirical Methods in Natural Language Processing}, pages 2241--2252.

\bibitem[{OpenAI(2024{\natexlab{a}})}]{GPT-4o-mini}
OpenAI. 2024{\natexlab{a}}.
\newblock \href {https://openai.com/index/gpt-4o-mini-advancing-cost-efficient-intelligence/} {Gpt-4o mini: advancing cost-efficient intelligence}.

\bibitem[{OpenAI(2024{\natexlab{b}})}]{o1-preview}
OpenAI. 2024{\natexlab{b}}.
\newblock \href {https://openai.com/index/learning-to-reason-with-llms/} {Learning to reason with llms}.

\bibitem[{OpenAI(2024{\natexlab{c}})}]{GPT-4o}
OpenAI. 2024{\natexlab{c}}.
\newblock \href {https://openai.com/index/hello-gpt-4o/} {Openai: Hello gpt-4o}.

\bibitem[{OpenAI(2024{\natexlab{d}})}]{o1-mini}
OpenAI. 2024{\natexlab{d}}.
\newblock \href {https://openai.com/index/openai-o1-mini-advancing-cost-efficient-reasoning/} {Openai o1-mini}.

\bibitem[{Pang et~al.(2022)Pang, Parrish, Joshi, Nangia, Phang, Chen, Padmakumar, Ma, Thompson, He et~al.}]{pang2022quality}
Richard~Yuanzhe Pang, Alicia Parrish, Nitish Joshi, Nikita Nangia, Jason Phang, Angelica Chen, Vishakh Padmakumar, Johnny Ma, Jana Thompson, He~He, et~al. 2022.
\newblock Quality: Question answering with long input texts, yes!
\newblock In \emph{Proceedings of the 2022 Conference of the North American Chapter of the Association for Computational Linguistics: Human Language Technologies}.

\bibitem[{Peng et~al.(2024)Peng, Quesnelle, Fan, and Shippole}]{peng2024yarn}
Bowen Peng, Jeffrey Quesnelle, Honglu Fan, and Enrico Shippole. 2024.
\newblock Yarn: Efficient context window extension of large language models.
\newblock In \emph{The Twelfth International Conference on Learning Representations}.

\bibitem[{Que et~al.(2024)Que, Duan, He, Mou, Zhou, Liu, Rong, Wang, Yang, Zhang et~al.}]{que2024hellobench}
Haoran Que, Feiyu Duan, Liqun He, Yutao Mou, Wangchunshu Zhou, Jiaheng Liu, Wenge Rong, Zekun~Moore Wang, Jian Yang, Ge~Zhang, et~al. 2024.
\newblock Hellobench: Evaluating long text generation capabilities of large language models.
\newblock \emph{arXiv preprint arXiv:2409.16191}.

\bibitem[{Reid et~al.(2024)Reid, Savinov, Teplyashin, Lepikhin, Lillicrap, Alayrac, Soricut, Lazaridou, Firat, Schrittwieser et~al.}]{reid2024gemini}
Machel Reid, Nikolay Savinov, Denis Teplyashin, Dmitry Lepikhin, Timothy Lillicrap, Jean-baptiste Alayrac, Radu Soricut, Angeliki Lazaridou, Orhan Firat, Julian Schrittwieser, et~al. 2024.
\newblock Gemini 1.5: Unlocking multimodal understanding across millions of tokens of context.
\newblock \emph{arXiv preprint arXiv:2403.05530}.

\bibitem[{Rein et~al.(2023)Rein, Hou, Stickland, Petty, Pang, Dirani, Michael, and Bowman}]{rein2023gpqa}
David Rein, Betty~Li Hou, Asa~Cooper Stickland, Jackson Petty, Richard~Yuanzhe Pang, Julien Dirani, Julian Michael, and Samuel~R Bowman. 2023.
\newblock Gpqa: A graduate-level google-proof q\&a benchmark.
\newblock \emph{arXiv preprint arXiv:2311.12022}.

\bibitem[{Shaham et~al.(2023)Shaham, Ivgi, Efrat, Berant, and Levy}]{shaham2023zeroscrolls}
Uri Shaham, Maor Ivgi, Avia Efrat, Jonathan Berant, and Omer Levy. 2023.
\newblock Zeroscrolls: A zero-shot benchmark for long text understanding.
\newblock In \emph{Findings of the Association for Computational Linguistics: EMNLP 2023}, pages 7977--7989.

\bibitem[{Song et~al.(2024)Song, Zheng, and Luo}]{song2024counting}
Mingyang Song, Mao Zheng, and Xuan Luo. 2024.
\newblock Counting-stars: A simple, efficient, and reasonable strategy for evaluating long-context large language models.
\newblock \emph{arXiv preprint arXiv:2403.11802}.

\bibitem[{Sprague et~al.(2024)Sprague, Yin, Rodriguez, Jiang, Wadhwa, Singhal, Zhao, Ye, Mahowald, and Durrett}]{sprague2024cot}
Zayne Sprague, Fangcong Yin, Juan~Diego Rodriguez, Dongwei Jiang, Manya Wadhwa, Prasann Singhal, Xinyu Zhao, Xi~Ye, Kyle Mahowald, and Greg Durrett. 2024.
\newblock To cot or not to cot? chain-of-thought helps mainly on math and symbolic reasoning.
\newblock \emph{arXiv preprint arXiv:2409.12183}.

\bibitem[{Tanzer et~al.(2024)Tanzer, Suzgun, Visser, Jurafsky, and Melas-Kyriazi}]{tanzerbenchmark}
Garrett Tanzer, Mirac Suzgun, Eline Visser, Dan Jurafsky, and Luke Melas-Kyriazi. 2024.
\newblock A benchmark for learning to translate a new language from one grammar book.
\newblock In \emph{The Twelfth International Conference on Learning Representations}.

\bibitem[{Team(2024)}]{qwen2.5}
Qwen Team. 2024.
\newblock \href {https://qwenlm.github.io/blog/qwen2.5/} {Qwen2.5: A party of foundation models}.

\bibitem[{Vodrahalli et~al.(2024)Vodrahalli, Ontanon, Tripuraneni, Xu, Jain, Shivanna, Hui, Dikkala, Kazemi, Fatemi et~al.}]{vodrahalli2024michelangelo}
Kiran Vodrahalli, Santiago Ontanon, Nilesh Tripuraneni, Kelvin Xu, Sanil Jain, Rakesh Shivanna, Jeffrey Hui, Nishanth Dikkala, Mehran Kazemi, Bahare Fatemi, et~al. 2024.
\newblock Michelangelo: Long context evaluations beyond haystacks via latent structure queries.
\newblock \emph{arXiv preprint arXiv:2409.12640}.

\bibitem[{Vrande{\v{c}}i{\'c} and Kr{\"o}tzsch(2014)}]{vrandevcic2014wikidata}
Denny Vrande{\v{c}}i{\'c} and Markus Kr{\"o}tzsch. 2014.
\newblock Wikidata: a free collaborative knowledgebase.
\newblock \emph{Communications of the ACM}, 57(10):78--85.

\bibitem[{Wang et~al.(2022)Wang, Pang, Chen, Phang, and Bowman}]{wang2022squality}
Alex Wang, Richard~Yuanzhe Pang, Angelica Chen, Jason Phang, and Samuel Bowman. 2022.
\newblock Squality: Building a long-document summarization dataset the hard way.
\newblock In \emph{Proceedings of the 2022 Conference on Empirical Methods in Natural Language Processing}, pages 1139--1156.

\bibitem[{Wang et~al.(2024{\natexlab{a}})Wang, Chen, Cheng, Liao, Zhang, Wu, Yu, Xu, Zhang, Luo et~al.}]{wang2024leave}
Minzheng Wang, Longze Chen, Fu~Cheng, Shengyi Liao, Xinghua Zhang, Bingli Wu, Haiyang Yu, Nan Xu, Lei Zhang, Run Luo, et~al. 2024{\natexlab{a}}.
\newblock Leave no document behind: Benchmarking long-context llms with extended multi-doc qa.
\newblock In \emph{Proceedings of the 2024 Conference on Empirical Methods in Natural Language Processing}, pages 5627--5646.

\bibitem[{Wang et~al.(2020)Wang, Wang, Han, Jiang, Han, Liu, Li, Li, Lin, and Zhou}]{wang2020maven}
Xiaozhi Wang, Ziqi Wang, Xu~Han, Wangyi Jiang, Rong Han, Zhiyuan Liu, Juanzi Li, Peng Li, Yankai Lin, and Jie Zhou. 2020.
\newblock Maven: A massive general domain event detection dataset.
\newblock In \emph{Proceedings of the 2020 Conference on Empirical Methods in Natural Language Processing (EMNLP)}, pages 1652--1671.

\bibitem[{Wang et~al.(2024{\natexlab{b}})Wang, Bukharin, Delalleau, Egert, Shen, Zeng, Kuchaiev, and Dong}]{wang2024helpsteer2}
Zhilin Wang, Alexander Bukharin, Olivier Delalleau, Daniel Egert, Gerald Shen, Jiaqi Zeng, Oleksii Kuchaiev, and Yi~Dong. 2024{\natexlab{b}}.
\newblock Helpsteer2-preference: Complementing ratings with preferences.
\newblock \emph{arXiv preprint arXiv:2410.01257}.

\bibitem[{Wei et~al.(2022)Wei, Wang, Schuurmans, Bosma, Xia, Chi, Le, Zhou et~al.}]{wei2022chain}
Jason Wei, Xuezhi Wang, Dale Schuurmans, Maarten Bosma, Fei Xia, Ed~Chi, Quoc~V Le, Denny Zhou, et~al. 2022.
\newblock Chain-of-thought prompting elicits reasoning in large language models.
\newblock \emph{Advances in neural information processing systems}, 35:24824--24837.

\bibitem[{Wu et~al.(2024{\natexlab{a}})Wu, Wang, Yu, Zhang, Chang, and Yu}]{wu2024longmemeval}
Di~Wu, Hongwei Wang, Wenhao Yu, Yuwei Zhang, Kai-Wei Chang, and Dong Yu. 2024{\natexlab{a}}.
\newblock Longmemeval: Benchmarking chat assistants on long-term interactive memory.
\newblock \emph{arXiv preprint arXiv:2410.10813}.

\bibitem[{Wu et~al.(2024{\natexlab{b}})Wu, Hee, Hu, and Lee}]{wu2024longgenbench}
Yuhao Wu, Ming~Shan Hee, Zhiqing Hu, and Roy Ka-Wei Lee. 2024{\natexlab{b}}.
\newblock Longgenbench: Benchmarking long-form generation in long context llms.
\newblock \emph{arXiv preprint arXiv:2409.02076}.

\bibitem[{Xiong et~al.(2024)Xiong, Liu, Molybog, Zhang, Bhargava, Hou, Martin, Rungta, Sankararaman, Oguz et~al.}]{xiong2024effective}
Wenhan Xiong, Jingyu Liu, Igor Molybog, Hejia Zhang, Prajjwal Bhargava, Rui Hou, Louis Martin, Rashi Rungta, Karthik~Abinav Sankararaman, Barlas Oguz, et~al. 2024.
\newblock Effective long-context scaling of foundation models.
\newblock In \emph{Proceedings of the 2024 Conference of the North American Chapter of the Association for Computational Linguistics: Human Language Technologies (Volume 1: Long Papers)}, pages 4643--4663.

\bibitem[{Xu et~al.(2024)Xu, Ye, Liu, Sun, Liu, Guo, Li, Liu, Huang, and Qiu}]{xu2024detectiveqa}
Zhe Xu, Jiasheng Ye, Xiangyang Liu, Tianxiang Sun, Xiaoran Liu, Qipeng Guo, Linlin Li, Qun Liu, Xuanjing Huang, and Xipeng Qiu. 2024.
\newblock Detectiveqa: Evaluating long-context reasoning on detective novels.
\newblock \emph{arXiv preprint arXiv:2409.02465}.

\bibitem[{Yao et~al.(2019)Yao, Ye, Li, Han, Lin, Liu, Liu, Huang, Zhou, and Sun}]{yao2019docred}
Yuan Yao, Deming Ye, Peng Li, Xu~Han, Yankai Lin, Zhenghao Liu, Zhiyuan Liu, Lixin Huang, Jie Zhou, and Maosong Sun. 2019.
\newblock Docred: A large-scale document-level relation extraction dataset.
\newblock In \emph{Proceedings of the 57th Annual Meeting of the Association for Computational Linguistics}, pages 764--777.

\bibitem[{Yao et~al.(2023)Yao, Chen, Lv, Cao, Xin, Yu, Jin, Xu, Zhang, Hou et~al.}]{yao2023viskop}
Zijun Yao, Yuanyong Chen, Xin Lv, Shulin Cao, Amy Xin, Jifan Yu, Hailong Jin, Jianjun Xu, Peng Zhang, Lei Hou, et~al. 2023.
\newblock Viskop: Visual knowledge oriented programming for interactive knowledge base question answering.
\newblock In \emph{Proceedings of the 61st Annual Meeting of the Association for Computational Linguistics (Volume 3: System Demonstrations)}, pages 179--189.

\bibitem[{Ye et~al.(2024)Ye, Wang, Huang, Chen, Zhang, Moniz, Gao, Geyer, Huang, Chen et~al.}]{ye2024justice}
Jiayi Ye, Yanbo Wang, Yue Huang, Dongping Chen, Qihui Zhang, Nuno Moniz, Tian Gao, Werner Geyer, Chao Huang, Pin-Yu Chen, et~al. 2024.
\newblock Justice or prejudice? quantifying biases in llm-as-a-judge.
\newblock \emph{arXiv preprint arXiv:2410.02736}.

\bibitem[{Yen et~al.(2024)Yen, Gao, Hou, Ding, Fleischer, Izsak, Wasserblat, and Chen}]{yen2024helmet}
Howard Yen, Tianyu Gao, Minmin Hou, Ke~Ding, Daniel Fleischer, Peter Izsak, Moshe Wasserblat, and Danqi Chen. 2024.
\newblock Helmet: How to evaluate long-context language models effectively and thoroughly.
\newblock \emph{arXiv preprint arXiv:2410.02694}.

\bibitem[{Zhang et~al.(2024{\natexlab{a}})Zhang, Liu, Lin, and Feng}]{zhang2024teaching}
Chen Zhang, Xiao Liu, Jiuheng Lin, and Yansong Feng. 2024{\natexlab{a}}.
\newblock Teaching large language models an unseen language on the fly.
\newblock \emph{arXiv preprint arXiv:2402.19167}.

\bibitem[{Zhang et~al.(2024{\natexlab{b}})Zhang, Bai, Lv, Gu, Liu, Zou, Cao, Hou, Dong, Feng et~al.}]{zhang2024longcite}
Jiajie Zhang, Yushi Bai, Xin Lv, Wanjun Gu, Danqing Liu, Minhao Zou, Shulin Cao, Lei Hou, Yuxiao Dong, Ling Feng, et~al. 2024{\natexlab{b}}.
\newblock Longcite: Enabling llms to generate fine-grained citations in long-context qa.
\newblock \emph{arXiv preprint arXiv:2409.02897}.

\bibitem[{Zhang et~al.(2024{\natexlab{c}})Zhang, Zhang, Ma, Li, Zhang, Li, Yao, Xu, Zhou, Zhang-Li et~al.}]{zhang2024tablellm}
Xiaokang Zhang, Jing Zhang, Zeyao Ma, Yang Li, Bohan Zhang, Guanlin Li, Zijun Yao, Kangli Xu, Jinchang Zhou, Daniel Zhang-Li, et~al. 2024{\natexlab{c}}.
\newblock Tablellm: Enabling tabular data manipulation by llms in real office usage scenarios.
\newblock \emph{arXiv preprint arXiv:2403.19318}.

\bibitem[{Zhang et~al.(2024{\natexlab{d}})Zhang, Chen, Hu, Xu, Chen, Hao, Han, Thai, Wang, Liu, and Sun}]{zhang2024infty}
Xinrong Zhang, Yingfa Chen, Shengding Hu, Zihang Xu, Junhao Chen, Moo Hao, Xu~Han, Zhen Thai, Shuo Wang, Zhiyuan Liu, and Maosong Sun. 2024{\natexlab{d}}.
\newblock \href {https://doi.org/10.18653/v1/2024.acl-long.814} {$\infty${B}ench: Extending long context evaluation beyond 100{K} tokens}.
\newblock In \emph{Proceedings of the 62nd Annual Meeting of the Association for Computational Linguistics (Volume 1: Long Papers)}, pages 15262--15277, Bangkok, Thailand. Association for Computational Linguistics.

\bibitem[{Zheng et~al.(2023)Zheng, Chiang, Sheng, Zhuang, Wu, Zhuang, Lin, Li, Li, Xing et~al.}]{zheng2023judging}
Lianmin Zheng, Wei-Lin Chiang, Ying Sheng, Siyuan Zhuang, Zhanghao Wu, Yonghao Zhuang, Zi~Lin, Zhuohan Li, Dacheng Li, Eric Xing, et~al. 2023.
\newblock Judging llm-as-a-judge with mt-bench and chatbot arena.
\newblock \emph{Advances in Neural Information Processing Systems}, 36:46595--46623.

\bibitem[{Zhong et~al.(2021)Zhong, Yin, Yu, Zaidi, Mutuma, Jha, Hassan, Celikyilmaz, Liu, Qiu et~al.}]{zhong2021qmsum}
Ming Zhong, Da~Yin, Tao Yu, Ahmad Zaidi, Mutethia Mutuma, Rahul Jha, Ahmed Hassan, Asli Celikyilmaz, Yang Liu, Xipeng Qiu, et~al. 2021.
\newblock Qmsum: A new benchmark for query-based multi-domain meeting summarization.
\newblock In \emph{Proceedings of the 2021 Conference of the North American Chapter of the Association for Computational Linguistics: Human Language Technologies}, pages 5905--5921.

\end{thebibliography}

\newpage
\appendix
\onecolumn

\section*{Appendix}

\section{Author Contributions}
\label{sec:ac}
\begin{itemize}[itemsep=0pt, leftmargin=*]
    \item Project lead: YB
    \item Benchmark design: YB, ST, JZ, HP, XW, XL, SC
    \item Annotation platform: ST, YB
    \item Annotator recruitment: YB, JX, ST, JZ
    \item Annotator management: YB, ST, JZ, HP, XL, SC
    \item Evaluation: YB, assisted by JZ, XL
    \item Writing: YB, ST, assisted by JZ, HP, XW, XL
    \item Supervision and fundraising: JL, LH, JT, YD
\end{itemize}

\section{Task Descriptions}
\label{sec:task}

\subsection*{\bluenote{I.1. Single-Document QA (Academic)}}
\textbf{\rednote{Task Description}}: Ask questions based on academic articles (papers, textbooks), excluding content related to charts and figures within the text.\\
\textbf{\rednote{Example Questions}}: 1. Which methods were used to collect data in the study? 2. In what ways does the author's argument align or conflict with the findings of Smith et al. (2020)?

\subsection*{\bluenote{I.2. Single-Document QA (Literary)}}
\textbf{\rednote{Task Description}}: Ask questions about literary works, potentially covering characters, plot, writing style, and central themes.\\
\textbf{\rednote{Example Questions}}: 1. What are the key traits that define [character]'s personality? 2. What is the turning point in the novel, and how does it impact the characters? 3. What message does the author seem to be conveying through the ending?

\subsection*{\bluenote{I.3. Single-Document QA (Legal)}}
\textbf{\rednote{Task Description}}: Ask questions based on legal documents, referencing scenarios like legal consultations, case analysis, or legal document review.\\
\textbf{\rednote{Example Questions}}: 1. What is the basis of the defendant's defense? 2. How is the estate distributed according to the will? 3. What are the conditions for tax incentives mentioned in this regulation?

\subsection*{\bluenote{I.4. Single-Document QA (Financial)}}
\textbf{\rednote{Task Description}}: Ask questions based on financial documents, including but not limited to financial report analysis, market analysis, investment strategies, and risk assessment.\\
\textbf{\rednote{Example Questions}}: 1. Based on the report, how do changes in operational expenses align with the company's revenue growth strategy? 2. What macroeconomic indicators are likely to impact the company's performance in the next fiscal year, and how are they addressed in the document? 3. How does the document evaluate the impact of regulatory changes on the company's capital structure?

\subsection*{\bluenote{I.5. Single-Document QA (Governmental)}}
\textbf{\rednote{Task Description}}: Ask questions based on government reports and official documents, potentially covering policies, regulations, and public facilities.\\
\textbf{\rednote{Example Questions}}: 1. What are the main allocations for healthcare in this year's government budget? 2. Who qualifies for the education grants mentioned in this document? 3. How does this policy address the concerns of small businesses?

\subsection*{\bluenote{I.6. Single-Document QA (Detective)}}
\textbf{\rednote{Task Description}}: Ask questions based on a detective or mystery novel. Questions must be inferable after reading most of the novel, such as who the murderer is or what the method of the crime was, without the full reasoning or answer being directly present in the text.\\
\textbf{\rednote{Example Questions}}: 1. Who murdered Mary?

\subsection*{\bluenote{I.7. Single-Document QA (Event ordering)}}
\textbf{\rednote{Task Description}}: Given a long text (usually a novel) and 4 plot events from the novel in random order, the model is required to select the correct sequence of the plot development.\\
\textbf{\rednote{Example Questions}}: 1. Order the four events in their original order...

\subsection*{\bluenote{II.1. Multi-Document QA (Academic)}}
\textbf{\rednote{Task Description}}: Ask questions based on academic articles (papers, textbooks), excluding content related to charts and figures. Questions must require using the information from at least 2 documents to be answered, with no irrelevant documents.\\
\textbf{\rednote{Example Questions}}: 1. What are the improvements of the method in paper A compared with paper B?

\subsection*{\bluenote{II.2. Multi-Document QA (Legal)}}
\textbf{\rednote{Task Description}}: Ask questions based on legal documents, requiring at least 2 documents. Questions must require information from each document to be answered, and there should be no irrelevant documents.\\
\textbf{\rednote{Example Questions}}: 1. Is Zhang's crime a case of imagined concurrence or statutory concurrence of crimes?

\subsection*{\bluenote{II.3. Multi-Document QA (Financial)}}
\textbf{\rednote{Task Description}}: Ask questions based on financial documents, requiring at least 2 documents. Questions must require information from each document to be answered, and there should be no irrelevant documents.\\
\textbf{\rednote{Example Questions}}: 1. How has the R\&D investment of the enterprises changed in the past ten years?

\subsection*{\bluenote{II.4. Multi-Document QA (Governmental)}}
\textbf{\rednote{Task Description}}: Ask questions based on government reports and official documents, requiring at least 2 documents. Questions must require information from each document to be answered, and there should be no irrelevant documents.\\
\textbf{\rednote{Example Questions}}: 1. How do the public transportation policies outlined in the 2022 Urban Development Report align with the environmental sustainability goals stated in the 2023 National Green Initiative document?

\subsection*{\bluenote{II.5. Multi-Document QA (Multi-news)}}
\textbf{\rednote{Task Description}}: Ask questions based on news articles, requiring at least 2 articles. Questions must require synthesizing information from multiple documents to be answered, and there should be no irrelevant documents.\\
\textbf{\rednote{Example Questions}}: 1. How have the top three positions in the medal leaderboard for the 2024 Paris Olympics changed over time?

\subsection*{\bluenote{III.1. Long In-context Learning (User guide QA)}}
\textbf{\rednote{Task Description}}: Given a long user guide, e.g., electronic device manual, software manual, musical instrument tutorial, annotate questions that require a deep understanding of the long text.\\
\textbf{\rednote{Example Questions}}: 1. I want to do time-lapse photography, how do I shoot it? 2. In what situations is it more effective to use parfor in MATLAB? 3. How can you change the timbre and achieve different expressive styles by controlling the force and speed of your key presses?

\subsection*{\bluenote{III.2. Long In-context Learning (New language translation)}}
\textbf{\rednote{Task Description}}: Translation tasks involving the rare languages Zhuang (vocabulary book and translation corpus from~\citet{zhang2024teaching}) and Kalamang (vocabulary book and translation corpus from~\citet{tanzerbenchmark}), requiring reading a vocabulary book to complete.\\
\textbf{\rednote{Example Questions}}: 1. Translate the following kalamang into English: Wa me kariak kaia kon untuk emumur kalo tumun amkeiret mu wara nanet.

\subsection*{\bluenote{III.3. Long In-context Learning (Many-shot learning)}}
\textbf{\rednote{Task Description}}: Given many-shot examples, answer the query based on the given examples. All label information is anonymized and can only be learned from the examples. This task primarily involves multi-class classification datasets, including the named entity recognition dataset FewNERD~\citep{ding2021few}, the relation classification dataset DocRED~\citep{yao2019docred}, the event detection dataset MAVEN~\citep{wang2020maven}, and the sentiment classification dataset GoEmotions~\citep{demszky2020goemotions}. \\
\textbf{\rednote{Example Questions}}: 1. What is the entity type of ``glucagon''? 2. What is the relation type between ``The Bone Forest'' and ``Robert Holdstock''? 3. What is the event type of ``became''? 4. What are the emotions of the document ``I'm more interested in why there are goldfish in the picture...''?

\subsection*{\bluenote{IV.1. Long-dialogue History Understanding (Agent history QA)}}
\textbf{\rednote{Task Description}}: Based on the agent dialogue history as context, ask questions about the content of the history. Specifically, we provide annotators with LLMs' dialogue history on playing games, which is derived from the GAMA-Bench~\cite{huang2024far}. This dataset includes eight classical multi-agent games categorized into three groups: Cooperative Games, Betraying Games, and Sequential Games. In our task, we use them as context and annotate questions for the agent interaction history. \\
\textbf{\rednote{Example Questions}}: 1. Which player is the most selfish one in the fourth round of the game?

\subsection*{\bluenote{IV.2. Long-dialogue History Understanding (Dialogue history QA)}}
\textbf{\rednote{Task Description}}: Given a multi-turn chat history between a user and an AI assistant, raise a question than demands understanding the dialogue history. To ensure the length of the history, we sample data from LongMemEval~\cite{wu2024longmemeval}, which consists of over 500 sessions for each chat history that challenges the long-term memory capabilities of LLMs. We take the chat history as context and raise new questions for long-dialogue understanding.  \\
\textbf{\rednote{Example Questions}}: 1. How long have I been living in my current apartment in Shinjuku?

\subsection*{\bluenote{V.1. Code Repository Understanding (Code repo QA)}}
\textbf{\rednote{Task Description}}: Based on a specific branch or commit of a codebase, annotate questions that require careful reading of multiple parts of the code or a deep understanding of the code's content to answer.\\
\textbf{\rednote{Example Questions}}: 1. For the current Megatron-LM framework, if I want to use the THD data format while enabling Context Parallel, how should I modify the experiments for rotary\_pos\_embedding?

\subsection*{\bluenote{VI.1. Long Structured Data Understanding (Table QA)}}
\textbf{\rednote{Task Description}}: Given a long table (e.g., financial report) or several interconnected tables, annotate questions that require integrating multiple cells or combining information from multiple tables. We provide annotators with long tables from the dataset proposed by TableLLM~\cite{zhang2024tablellm}. \\
\textbf{\rednote{Example Questions}}: 1. For the industry fields involving entertainment, which grows most largely from 2021 to 2023?

\subsection*{\bluenote{VI.2. Long Structured Data Understanding (Knowledge graph reasoning)}}
\textbf{\rednote{Task Description}}: Given a large-scale knowledge graph, annotate questions and corresponding answers that require integrating multiple entities. We construct the knowledge graph (extracted from Wikidata~\cite{vrandevcic2014wikidata}) and the complex logical queries based on the KQAPro dataset~\cite{cao2022kqa}. Groundtruth answers are automatically derived by running the corresponding KoPL program~\cite{cao2022kqa,yao2023viskop} on the graph. \\
\textbf{\rednote{Example Questions}}: 1. When did the people who captured Q10549 come to the region where Q231 is located?

\section{Annotation Details}

\subsection{Annotation Platform}
\label{sec:platform}

Our annotation platform includes three pages: main page, data annotation page, and data verification page.

\begin{figure}[t]
    \centering
    \includegraphics[width=\linewidth]{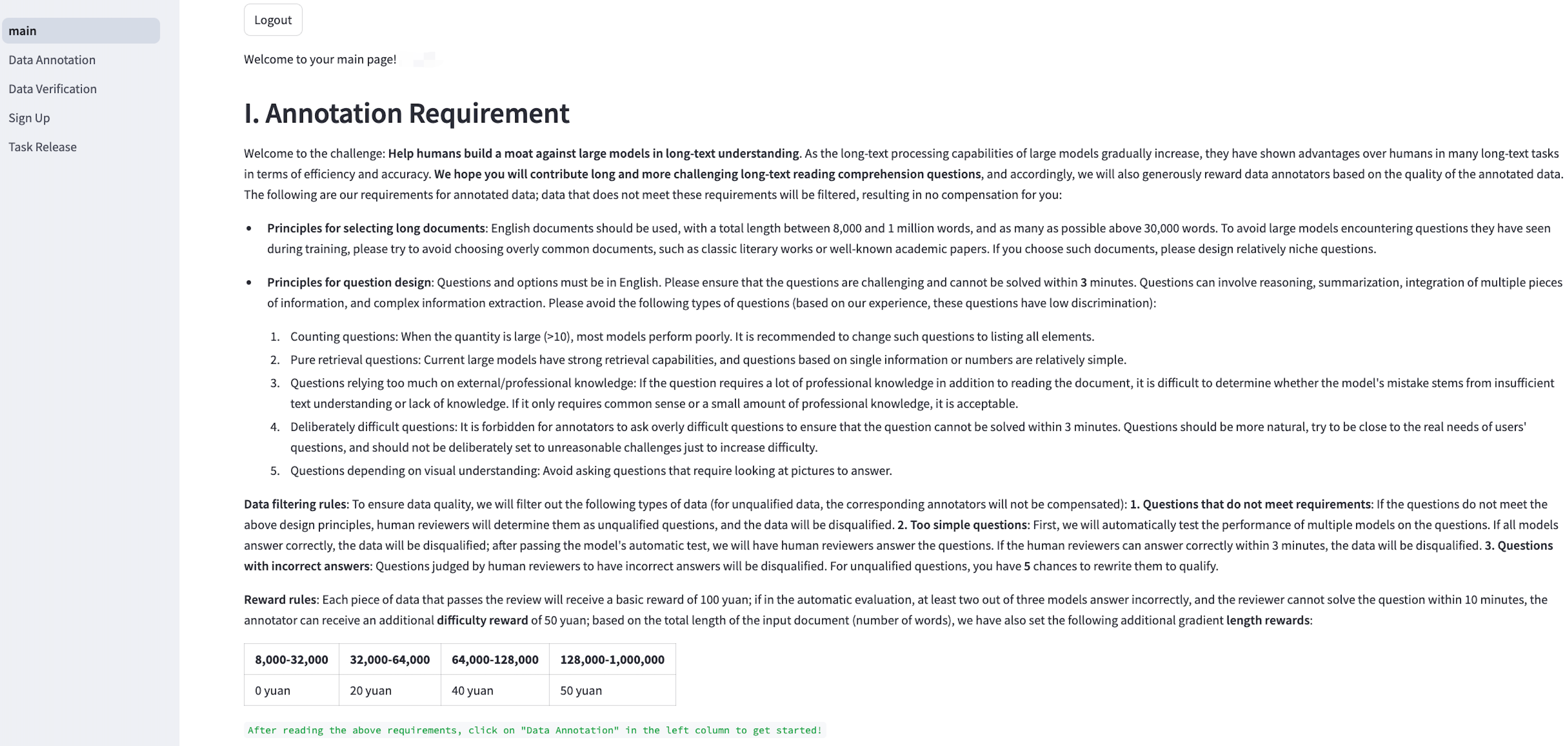}
    \caption{Screenshot of the \rednote{main page} (top part). After logging in, the annotator will first see this page, which displays our requirements and incentive policies. Annotators can also see the statuses of their data on this page.}
    \label{fig:Main}
\end{figure}

\begin{figure*}[htbp]
    \centering
    \includegraphics[width=\linewidth]{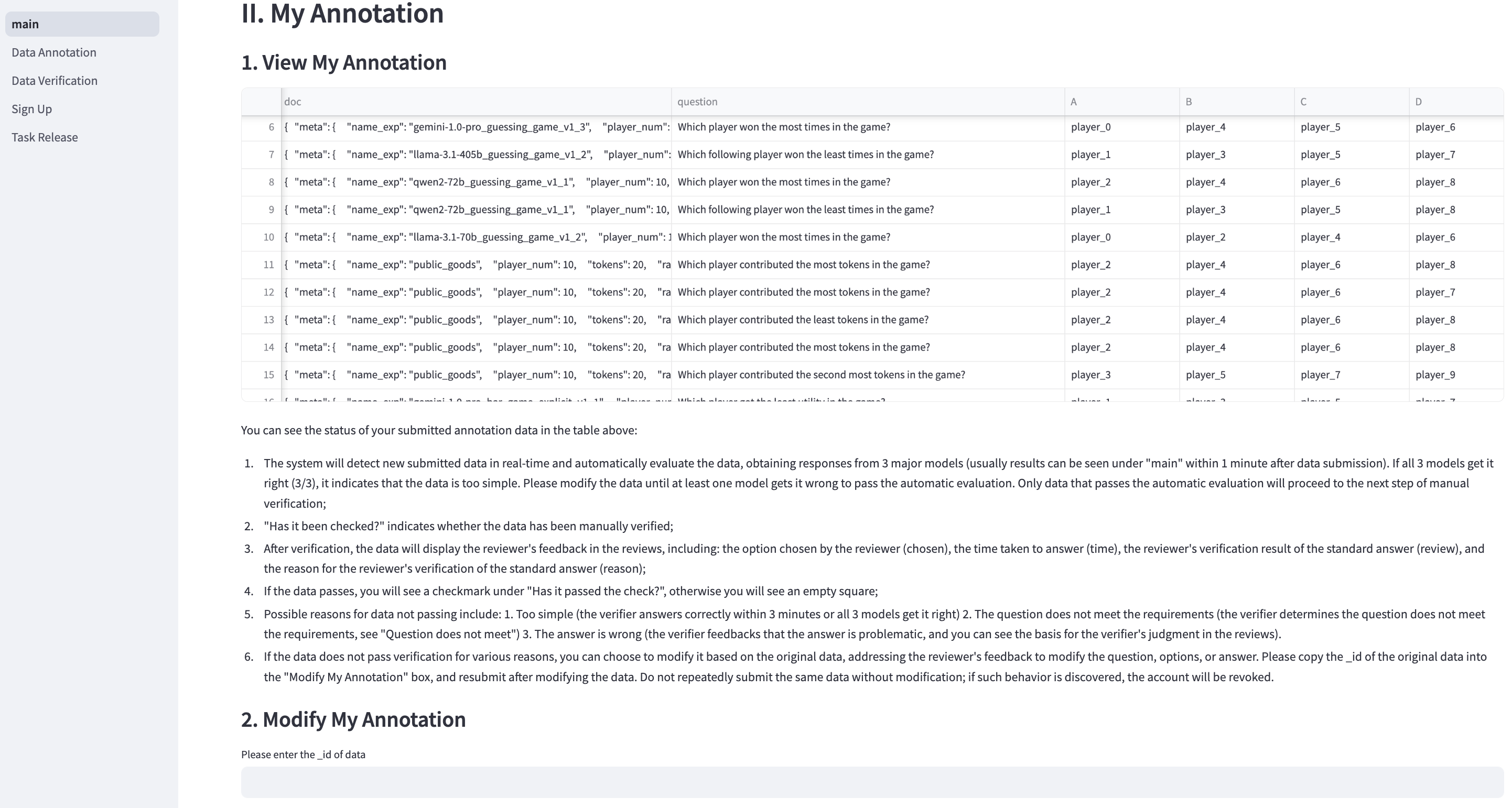}
    \caption{Screenshot of the \rednote{main page} (bottom part). Annotators can view the status of their data on this page. They can modify their rejected data for resubmission.}
    \label{fig:Main2}
\end{figure*}

\noindent
\textbf{\rednote{Main page.}}
The main page serves as the central hub of the website, providing an overview of the tasks and data. Figure~\ref{fig:Main} shows the top part of the main page, where we display the annotation requirements for our task, allowing users to understand the demand of our annotation task. The bottom part of the main page, as shown in Figure~\ref{fig:Main2}, also includes functionality to view the data status, where the feedback from automated and manual reviews is displayed. It also handles the deletion and modification of data. Each user can only view their own data and is not able to access others.

\begin{figure}[htbp]
    \centering
    \includegraphics[width=\linewidth]{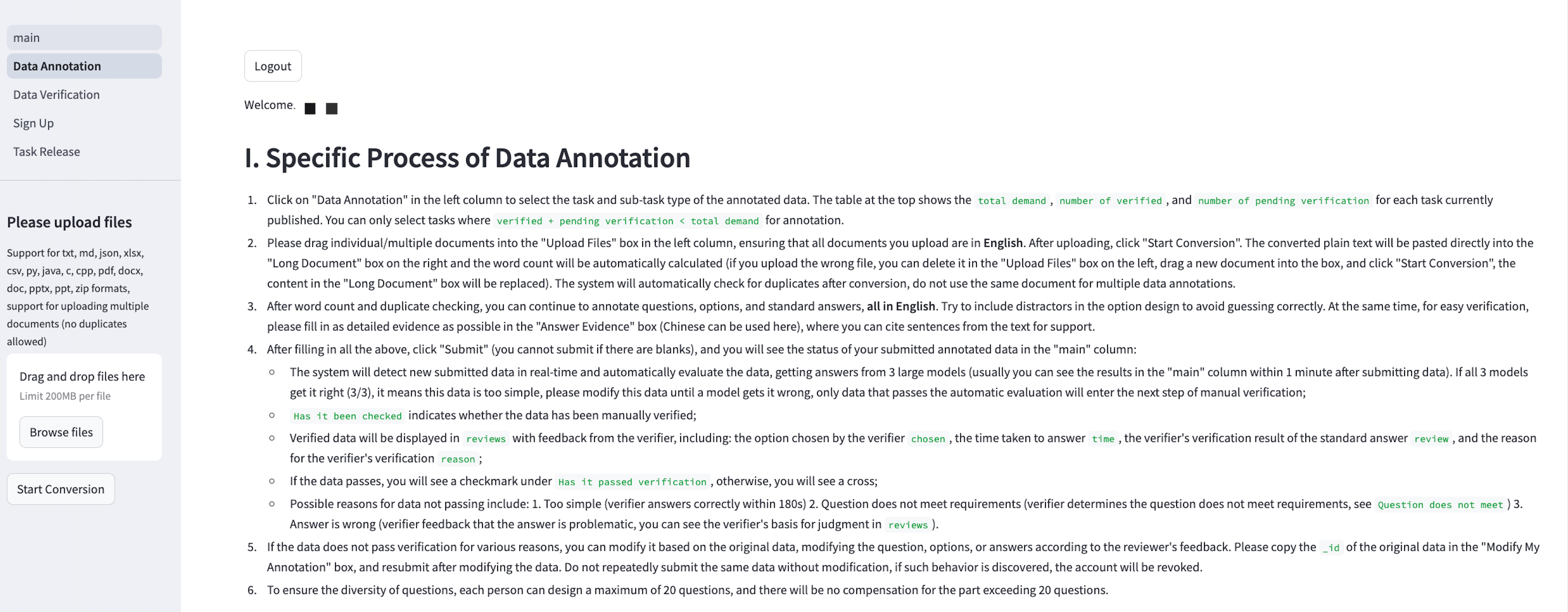}
    \caption{Screenshot of the \rednote{data annotation page} (top part).}
    \label{fig:Data_Annotation}
\end{figure}

\begin{figure}[htbp]
    \centering
    \includegraphics[width=0.9\linewidth]{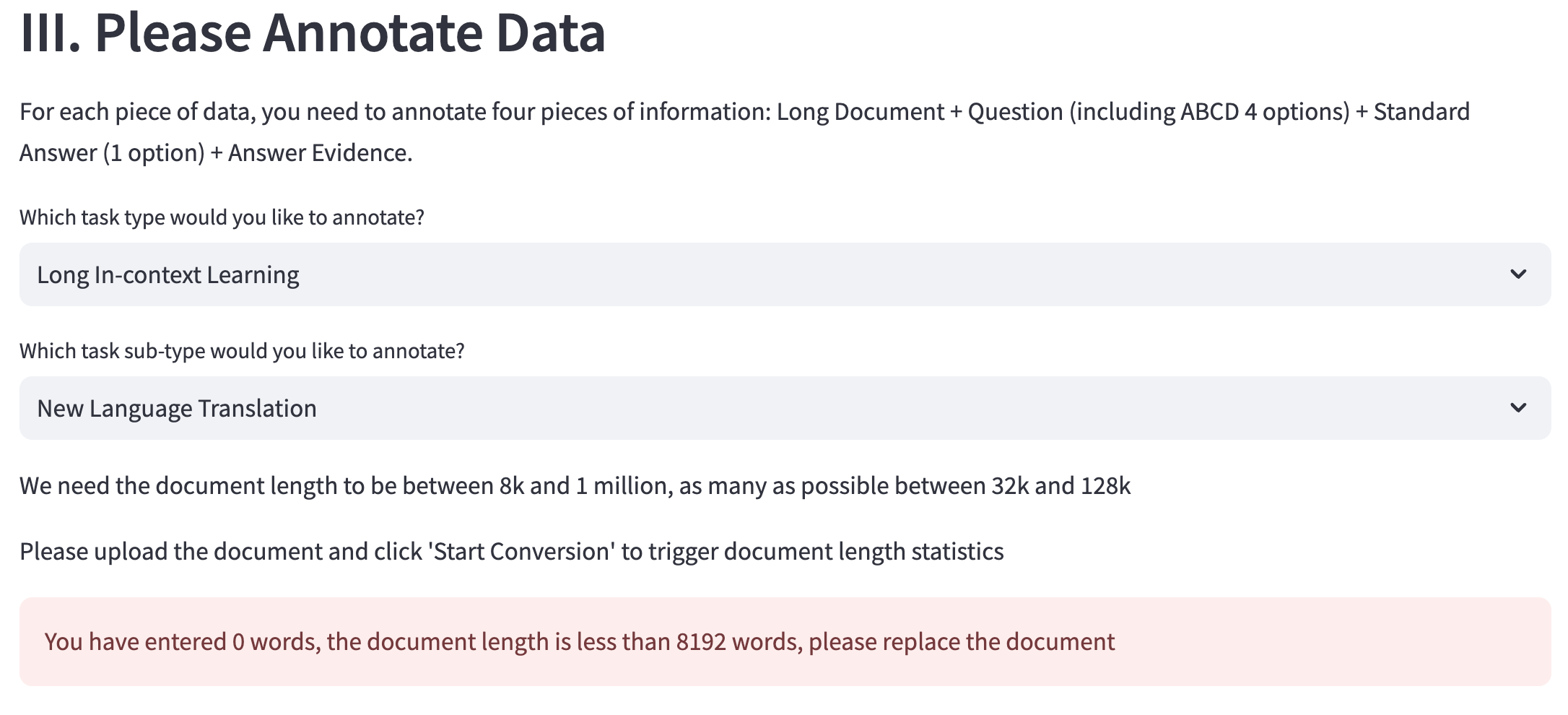}
    \caption{Screenshot of the \rednote{data annotation page} (bottom part). Annotator first uploads the document(s) and proposes a multiple-choice question based on the content.}
    \label{fig:Data_Annotation2}
\end{figure}

\noindent
\textbf{\rednote{Data annotation page.}} This page is designed for users to annotate long-context QA data. As shown in Figure~\ref{fig:Data_Annotation}, our guideline instructs users through the process of selecting tasks and subtasks, uploading documents, and annotating questions, options, and answers. The page ensures that all annotations are in English and meet specific requirements to challenge LLMs. As shown in Figure~\ref{fig:Data_Annotation2}, annotators will first choose the task category they would like to annotate, then upload their documents to annotate a multiple-choice question. Our platform includes features to check for the word count and duplicate documents to ensure the length and diversity of documents. After questions are annotated, we conduct automated reviews to verify the complexity of the questions to ensure they are not overly simple. The page also provides instructions for annotating data and limits the number of questions each user can annotate to maintain diversity. 

\begin{figure}[htbp]
    \centering
    \includegraphics[width=\linewidth]{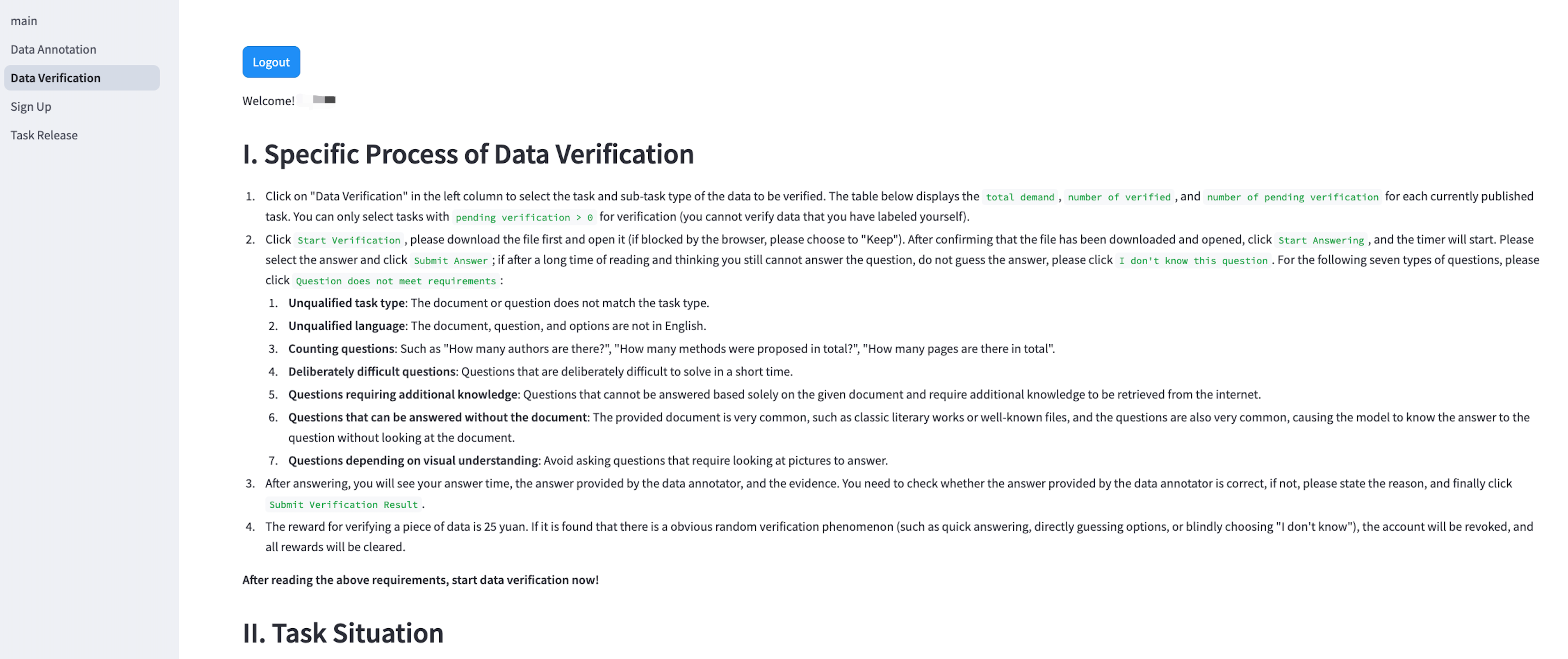}
    \caption{Screenshot of the \rednote{data verification page} (requirements part). Manual review will be conducted on this page to check whether the annotated data aligns with our requirements.}
    \label{fig:Data_Verification}
\end{figure}

\begin{figure}[htbp]
    \centering
    \includegraphics[width=\linewidth]{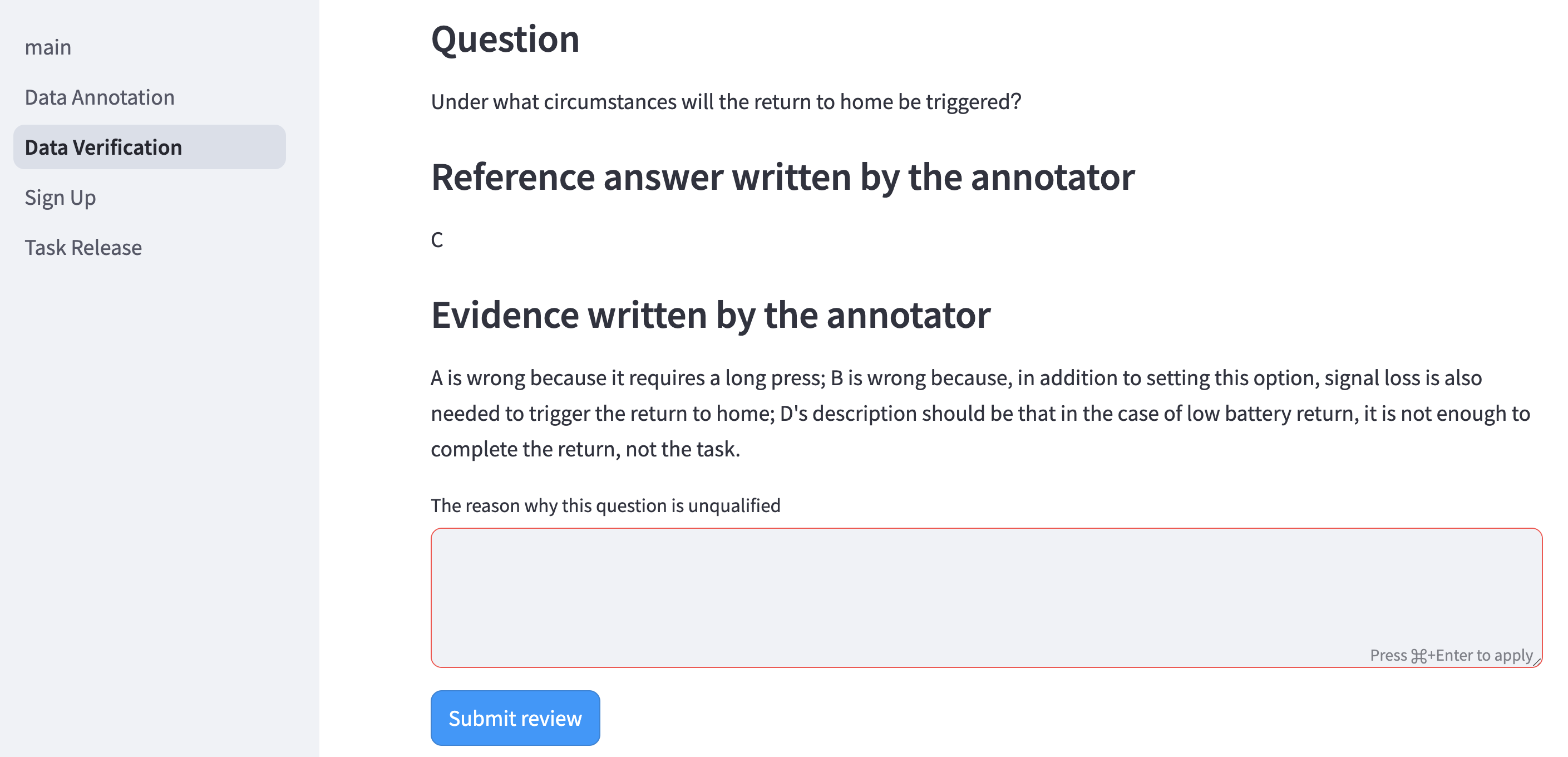}
    \caption{Screenshot of the \rednote{data verification page} after clicking the ``Question does not meet requirements'' button. Reviewers will use this page to write rejecting reasons if they decide that this question is unqualified.}
    \label{fig:Data_Verification3}
\end{figure}

\begin{figure}[htbp]
    \centering
    \includegraphics[width=\linewidth]{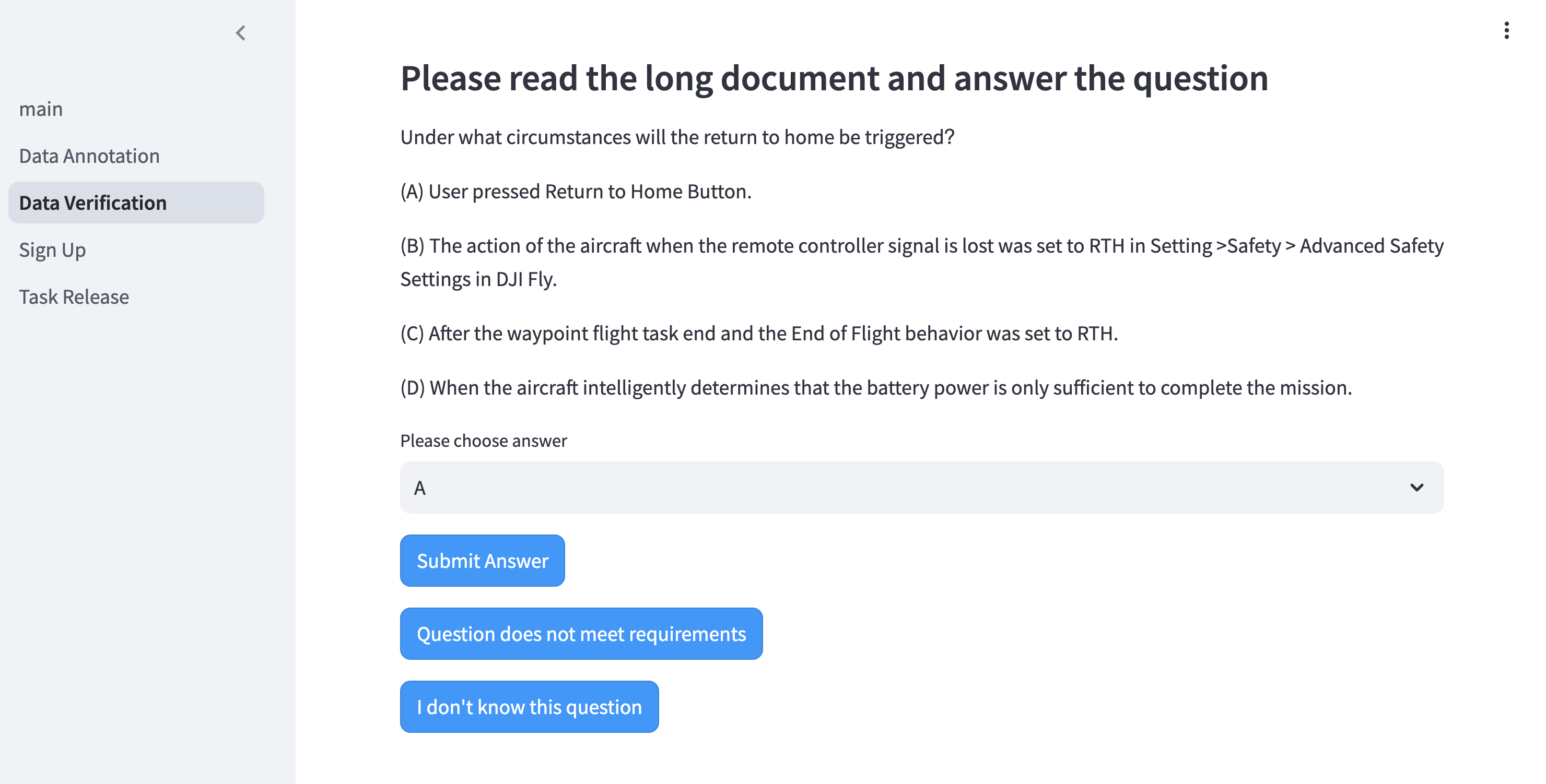}
    \caption{Screenshot of the \rednote{data verification page} for solving the question. Reviewers will enter this page when they attempt to answer the question. The long documents were downloaded before they answer the question.}
    \label{fig:Data_Verification2}
\end{figure}

\begin{figure}[htbp]
    \centering
    \includegraphics[width=\linewidth]{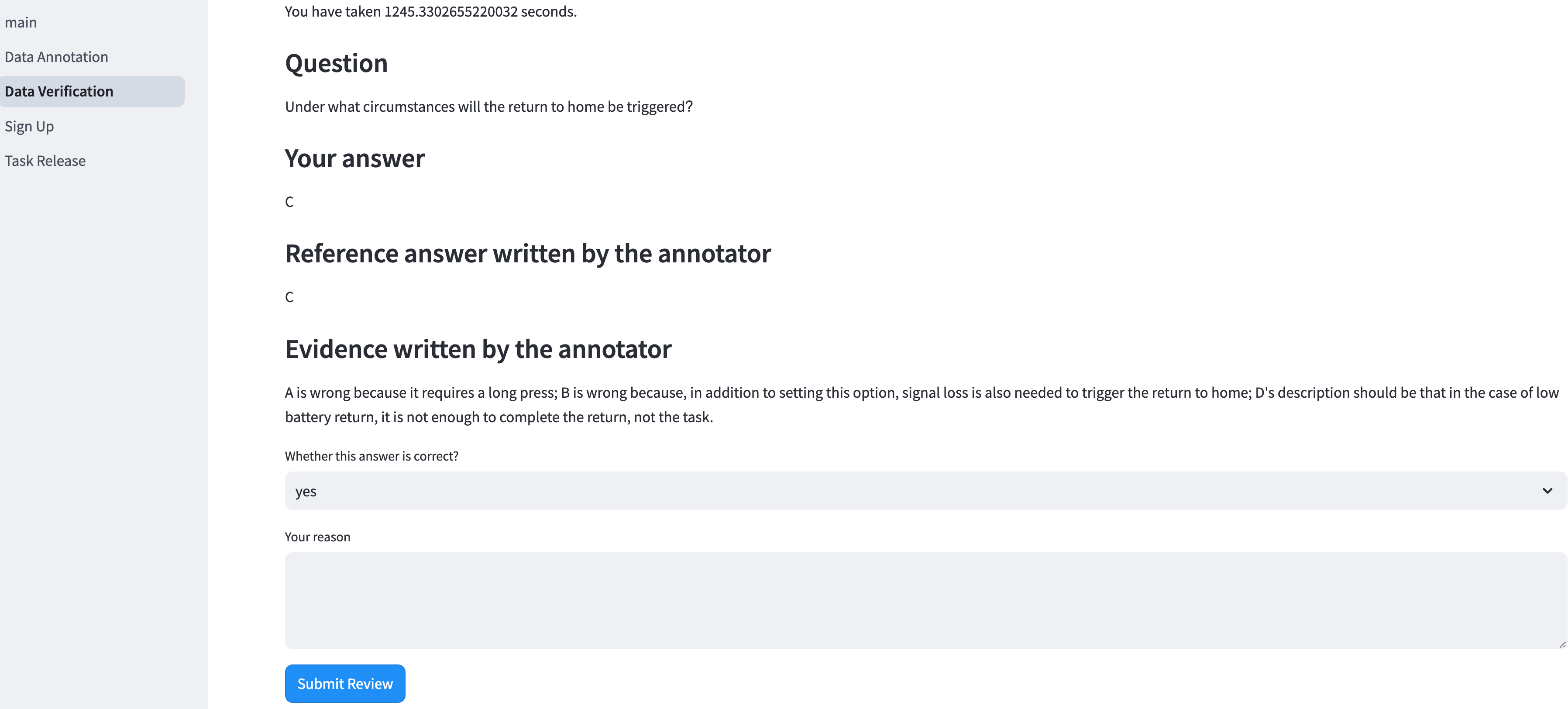}
    \caption{Screenshot of the \rednote{data verification page} after clicking the ``Submit Answer'' button. Reviewers will use this page to check whether the reference answer is correct and submit their reason.}
    \label{fig:Data_Verification4}
\end{figure}

\noindent
\textbf{\rednote{Data verification page.}}
As illustrated in Figure~\ref{fig:Data_Verification}, the data verification page is where human experts review the annotated data for accuracy and quality. Reviewers can only verify data that has passed the automated review and cannot verify data annotated by themselves. The page requires reviewers to download the documents and submit their own choice, and provide feedback on the correctness of the groundtruth answers. As shown in Figure~\ref{fig:Data_Verification3}, this page also allows users to flag questions that do not meet the requirements, such as those that do not match the task type, or require additional knowledge beyond the provided document. If the question is qualified, then the reviewer will attempt to answer it, as shown in Figure~\ref{fig:Data_Verification2}. This process includes a timer to track the time taken to answer each question. Figure~\ref{fig:Data_Verification4} shows the page when the reviewer finishes answering the question. The reviewer will be able to read the answer and evidence written by the annotator. The reviewer may check whether the answer is correct and submit the reason.

\subsection{Annotator Statistics}
\label{sec:stat}

\begin{figure}[htbp]
    \centering
    \subfigure[Age]{
        \includegraphics[width=0.26\textwidth]{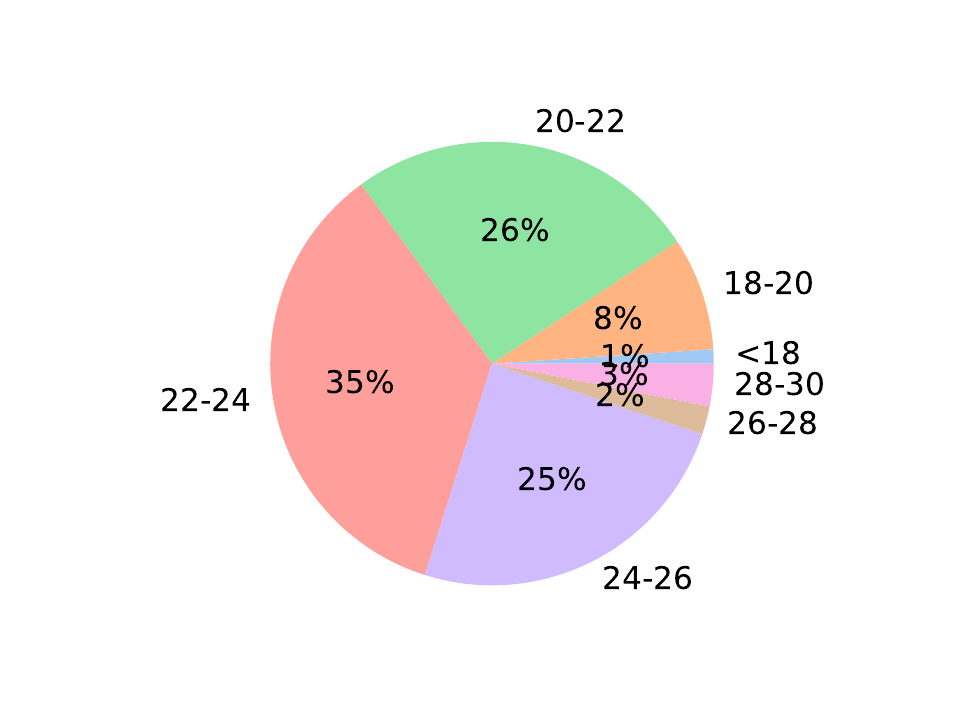}
    }
    \subfigure[Gender]{
        \includegraphics[width=0.17\textwidth]{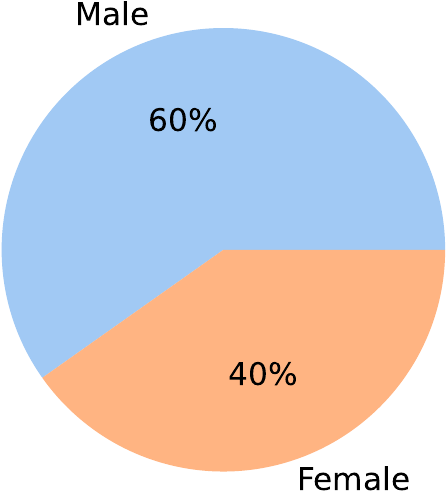}
    }
    \subfigure[Major]{
        \includegraphics[width=0.26\textwidth]{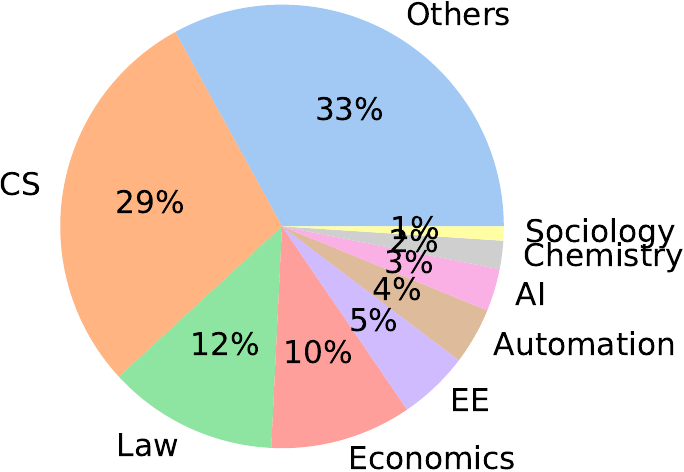}
    }
    \subfigure[Degree]{
        \includegraphics[width=0.20\textwidth]{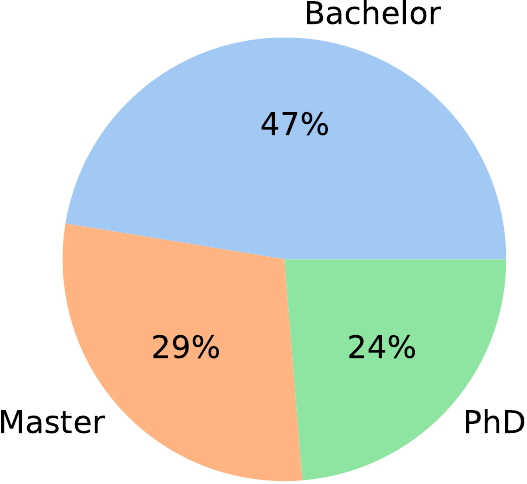}
    }
    \caption{Distribution of our annotators across ages, genders, majors, and degrees.}
    \label{fig:all_stats}
\end{figure}

To understand how diverse and professional our annotators are, we ask our annotators to fill in their age, gender, major, and degree during registration. We have ensured that no personal privacy information is leaked.
Figure~\ref{fig:all_stats} displays the diverse distribution of annotators across various dimensions. In terms of age, the majority of annotators fall within the 20-22 (26\%), 22-24 (35\%), and 24-26 (25\%) age groups because almost all annotators are recruited from universities. The distribution of majors is sufficiently diverse, with Computer Science (CS) being the most common (29\%), followed by Law (24\%) and Economics (22\%). Finally, the majority of annotators are holding or pursuing a Bachelor's degree (47\%), with a smaller proportion holding a Master's (29\%) or PhD (24\%). Each annotator can annotate at most 20 data to ensure the diversity of the data.

\subsection{Annotation Guidelines}
\label{sec:guide}

\noindent
\textbf{\bluenote{\underline{Overall annotation and platform guideline, displayed on the main page:}}}

\begin{tcolorbox}[size=title,opacityfill=0.05,breakable]
\noindent
Welcome to the challenge: \textbf{Help humans build a moat against AI systems in long-context understanding}. As the long-context processing capabilities of large language models gradually increase, they have shown advantages over humans in many long-context tasks in terms of efficiency and accuracy. We invite you to contribute long and challenging long-context reading comprehension questions, and accordingly, we will also generously reward data annotators based on the quality of the annotated data. The following are our requirements for annotated data; data that does not meet these requirements will be filtered, resulting in no payment:
\\ \\
- \textbf{Principles for selecting long documents}: English documents should be used, with a total length between 8,000 and 2 million words, and as many as possible above 32,000 words. To avoid large language models encountering questions they have seen during training, please try to avoid choosing overly common documents, such as classic literary works or well-known academic papers. If you choose such documents, please design relatively niche questions.
\\ \\
- \textbf{Principles for question design}: Questions and options must be in English. Please make sure that the questions are challenging enough and cannot be solved within \textbf{3} minutes. Questions can involve reasoning, summarization, integration of multiple pieces of information, and complex information extraction. Please avoid the following types of questions (based on our experience, these questions have low discrimination):

1. \emph{Counting-type questions}: When the quantity is large (>10), most models perform poorly. It is recommended to change such questions to listing all elements.

2. \emph{Retrieval-type questions}: Current large language models have strong retrieval capabilities, and questions based on single information located somewhere in the document are relatively simple.

3. \emph{Questions that rely too much on external/professional knowledge}: If the question requires a lot of professional knowledge in addition to reading the document, it is difficult to determine whether the model's mistake stems from insufficient text understanding or lack of knowledge. It is acceptable if it only requires common sense or a small amount of professional knowledge.

4. \emph{Deliberately difficult questions}: It is forbidden for annotators to ask deliberately difficult and stilted questions just to ensure that the human reviewer cannot solve them within a short amount of time. Questions should be more natural, try to be close to the real needs of users' questions, and should not be deliberately set to unreasonable challenges just to increase difficulty.

5. \emph{Questions that depend on visual understanding}: Avoid asking questions that require looking at pictures to answer.
\\ \\
\textbf{Data filtering rules}: To ensure data quality, we will filter out the following types of data (for unqualified data, the corresponding annotators will not be rewarded, and you have 5 chances to rewrite them to qualify): 

1. \emph{Questions that do not meet requirements}: If the questions do not meet the above requirements, human reviewers will determine them as unqualified questions, and the data will be disqualified. 

2. \emph{Too simple questions}: First, we will automatically test the performance of three models on the questions. If all models answer correctly, the data will be disqualified; after passing the model's automatic test, we will have human reviewers answer the questions. If the human reviewers can answer correctly within 3 minutes, the data will be disqualified. 

3. \emph{Questions with incorrect answers}: Questions judged by human reviewers to have incorrect answers will be disqualified. 
\\ \\
\textbf{Reward rules}: Each piece of data that passes the review will receive a basic reward of \texttt{100} \texttt{CNY}; if in the automatic evaluation, at least two out of three models answer incorrectly, and the reviewer cannot solve the question within 10 minutes, the annotator can receive an additional \textbf{difficulty reward} of \texttt{50} \texttt{CNY}; based on the total length of the input document (number of words), we have also set the following additional stepped \textbf{length rewards}:

8,000 - 32,000 words: \texttt{0} \texttt{CNY}

32,000 - 64,000 words: \texttt{20} \texttt{CNY}

64,000 - 128,000 words: \texttt{40} \texttt{CNY}

128,000 - 1,000,000 words: \texttt{50} \texttt{CNY}
\\ \\
After reading the above requirements, click on ``Data Annotation'' in the left column to get started!

\end{tcolorbox}

\noindent
\textbf{\bluenote{\underline{Guidelines provided to the annotators, displayed on the data annotation page:}}}

\begin{tcolorbox}[size=title,opacityfill=0.05,breakable]
\noindent
1. Click on ``Data Annotation'' in the left column to select the task and subtask type of the annotated data. The table at the top shows the ``total demand'', ``number of verified'', and ``number of pending verification'' for each task. You can only select tasks where ``verified + pending verification < total demand'' for annotation.

2. Please drag individual/multiple files into the ``Upload Files'' box in the left column. Make sure that all files you upload are in \textbf{English}. After uploading, click ``Start Conversion''. The converted plain text will be pasted directly into the ``Long Document'' box on the right and the word count will be automatically calculated. If you upload the wrong file, you can delete it in the ``Upload Files'' box on the left, drag a new document into the box, and click ``Start Conversion'', the content in the ``Long Document'' box will be replaced. The system will automatically check for duplicates after conversion, do not use the same document for multiple submissions.
    
3. After passing word counting and duplicate checking, you can continue to annotate questions, options, and answers, all in English. Try to include distractors in the option design to avoid guessing correctly. At the same time, for ease of verification, please fill in as detailed evidence as possible in the ``Evidence'' box, where you can cite sentences from the long context for support.
    
4. After filling in all the above, click ``Submit'' (you cannot submit if there are blanks), and you will see the status of your submitted annotated data in the ``main'' column:

- The system will detect newly submitted data in real-time and automatically evaluate the data, getting answers from 3 large language models (usually you can see the results in the ``main'' column within 1 minute after submitting data). If all 3 models get it right (3/3), it means this data is too simple, please modify this data until at least one model gets it wrong, only data that passes the automatic evaluation will enter the next step of manual verification.

- ``\texttt{Checked?}'' indicates whether the data has been manually verified.

- Verified data will be displayed in ``\texttt{reviews}'' with feedback from the verifier, including the option chosen by the verifier (``\texttt{chosen}''), the time taken to answer (``\texttt{time}''), the verifier's verification result of the groundtruth answer (``\texttt{correctness}''), and the reason for the verifier's judgment (``\texttt{reason}'').

- If the data passes, you will see a checkmark under ``\texttt{Verification passed?}'', otherwise, you will see a cross.

- Possible reasons for data not passing include: (1). Too simple (3/3 models get it right or verifier answers correctly within 180s); (2). Question does not meet requirements (verifier determines the question does not meet requirements, see the ``\texttt{reason}'' box for the detailed reason); (3). The answer is wrong (you can see the verifier's basis for judgment in ``\texttt{reason}'').

5. If the data does not pass verification for various reasons, you can modify it based on the original data, modifying the question, options, or answer according to the reviewer's feedback. Please copy the ``\texttt{\_id}'' of the original data in the ``\texttt{Modify My Annotation}'' box, and resubmit after modifying the data. Do not repeatedly submit the same data without modification, if such behavior is discovered, the account will be revoked.
    
6. To ensure the diversity of questions, each user can design a maximum of 20 questions.

\end{tcolorbox}

\noindent
\textbf{\bluenote{\underline{Guidelines for the reviewers, displayed on the data verification page:}}}

\begin{tcolorbox}[size=title,opacityfill=0.05,breakable]
\noindent
1. Click on ``Data Verification'' in the left column to select the task and subtask type of the data to be verified. The table below displays the ``total demand'', ``number of verified'', and ``number of pending verification'' for each current task. You can only select tasks with ``pending verification > 0'' for verification (you cannot verify data that you have labeled yourself).

2. Click ``Start Verification'', please download the file first and open it (if blocked by the browser, please choose ``Keep''). After confirming that the file has been downloaded and opened, click ``Start Answering'', and the timer will start. Please select the answer and click ``Submit Answer''; if after a long time (>15 min) of reading and thinking you still cannot answer the question, do not guess the answer, please click ``I don't know the answer''. For the following seven types of questions, please click ``Question does not meet requirements'':
(1) \emph{Mismatched task type}: The document or question does not match the task type.
(2) \emph{Unqualified language}: The document, question, and options are not in English.
(3) \emph{Counting questions}: Such as ``How many authors are there?'', ``How many methods were proposed in total?'', ``How many pages are there in total''.
(4) \emph{Deliberately difficult questions}: Questions that are deliberately difficult to solve in a short time.
(5) \emph{Questions requiring additional knowledge}: Questions that cannot be answered based solely on the given document and require additional knowledge to be searched from the internet.
(6) \emph{Questions that can be answered without the document}: The provided document is very common, such as classic literary works or well-known files, and the questions are also very common, causing the model to know the answer to the question without looking at the document.
(7) \emph{Questions depending on visual understanding}: Questions that require looking at visual contents to answer.

3. After answering, you will see your answer time, the answer provided by the data annotator, and the evidence. You need to check whether the answer provided by the data annotator is correct, if not, please fill in the reason, and finally click ``Submit Verification Result''.

4. The reward for verifying a piece of data is \texttt{25} \texttt{CNY}. If it is found that there is a malicious verification pattern (such as quick answering, directly guessing options, or blindly choosing ``I don't know the answer''), the account will be revoked, and all rewards will be cleared.

\textbf{After reading the above requirements, start data verification now!}

\end{tcolorbox}

\subsection{Data Collection Cost}
We spend approximately \texttt{100,000 CNY} on data collection.

\section{More Evaluation Details}
\label{sec:setup}

\subsection{Baseline Models}
Our open-source baselines include: GLM-4-9B-Chat~\cite{glm2024chatglm}, Llama-3.1-8B-Instruct, Llama-3.1-70B-Instruct, Llama-3.3-70B-Instruct~\cite{dubey2024llama}, Llama-3.1-Nemotron-70B-Instruct~\cite{wang2024helpsteer2}, Qwen2.5-7B-Instruct, Qwen2.5-72B-Instruct~\cite{qwen2.5}, Mistral-Large-Instruct-2407, Mistral-Large-Instruct-2411~\cite{jiang2023mistral}, and c4ai-command-r-plus-08-2024~\cite{cohere_for_ai_2024}.
Our proprietary baselines include: \href{https://open.bigmodel.cn/pricing}{GLM-4-Plus}~\cite{glm2024chatglm}, GPT-4o-mini-2024-07-18~\cite{GPT-4o-mini}, GPT-4o-2024-08-06, GPT-4o-2024-11-20~\cite{GPT-4o}, o1-mini-2024-09-12~\cite{o1-mini}, o1-preview-2024-09-12~\cite{o1-preview}, and Claude-3.5-Sonnet-20241022~\cite{claude-3-5}.
All of the models mentioned above have a context window length of 128k tokens, with the exception of Claude-3.5-Sonnet-20241022, which has a context window length of 200k tokens.

\subsection{Evaluation Setting}
In the zero-shot evaluation setting, we set the generation sampling parameters to \texttt{temperature}=0.1 and \texttt{max\_new\_tokens}=128. In the zero-shot + CoT setting, for the first model call where the model generates the chain-of-thought, we set \texttt{temperature}=0.1 and \texttt{max\_new\_tokens}=1024. For the subsequent model call where the model outputs the final answer, we set \texttt{temperature}=0.1 and \texttt{max\_new\_tokens}=128.

\subsection{Evaluation Prompts}

\xhdr{Prompt for zero-shot setting}

\begin{tcolorbox}[size=title,opacityfill=0.05,breakable]
\noindent
Please read the following text and answer the question below.
\\ \\
<text>

\{\emph{Long Context}\}

</text>
\\ \\
What is the correct answer to this question: \{\emph{Question}\}

Choices:

(A) \{\emph{Choice A}\}

(B) \{\emph{Choice B}\}

(C) \{\emph{Choice C}\}

(D) \{\emph{Choice D}\}
\\ \\
Format your response as follows: ``The correct answer is (insert answer here)''.
\end{tcolorbox}

\xhdr{Prompt for zero-shot + CoT setting}

\begin{tcolorbox}[size=title,opacityfill=0.05,breakable]
\noindent
Please read the following text and answer the question below.
\\ \\
<text>

\{\emph{Long Context}\}

</text>
\\ \\
What is the correct answer to this question: \{\emph{Question}\}

Choices:

(A) \{\emph{Choice A}\}

(B) \{\emph{Choice B}\}

(C) \{\emph{Choice C}\}

(D) \{\emph{Choice D}\}
\\ \\
Let’s think step by step:
\end{tcolorbox}

\begin{tcolorbox}[size=title,opacityfill=0.05,breakable]
\noindent
Please read the following text and answer the questions below.
\\ \\
The text is too long and omitted here.
\\ \\
What is the correct answer to this question: \{\emph{Question}\}

Choices:

(A) \{\emph{Choice A}\}

(B) \{\emph{Choice B}\}

(C) \{\emph{Choice C}\}

(D) \{\emph{Choice D}\}
\\ \\
Let’s think step by step: \{\emph{Chain of thought generated in the last response}\}
\\ \\
Based on the above, what is the single, most likely answer choice? Format your response as follows: ``The correct answer is (insert answer here)''.
\end{tcolorbox}

\section{Deferred Experimental Results}

\begin{table}[htbp]
\centering
\resizebox{\linewidth}{!}{
\begin{tabular}{l|cc|cc|cc|cc|cc|cc}
\toprule
 &  & & \multicolumn{4}{|c}{\textbf{Difficulty}} & \multicolumn{6}{|c}{\textbf{Length (<32k; 32k-128k; >128k)}} \\
\cmidrule(r){1-3} \cmidrule(lr){4-7} \cmidrule(l){8-13}
\textbf{Model} & \multicolumn{2}{c|}{\textbf{Overall}} & \multicolumn{2}{c|}{\textbf{Easy}} & \multicolumn{2}{c|}{\textbf{Hard}} & \multicolumn{2}{c|}{\textbf{Short}} & \multicolumn{2}{c|}{\textbf{Medium}} & \multicolumn{2}{c}{\textbf{Long}} \\ 
\midrule
\texttt{Qwen2.5-7B-Instruct} & 27.0 & \cellcolor{mygray}29.8 & 29.2 & \cellcolor{mygray}30.7 & 25.7 & \cellcolor{mygray}29.3 & 36.1 & \cellcolor{mygray}35.6 & 23.7 & \cellcolor{mygray}26.5 & 18.5 & \cellcolor{mygray}26.9 \\
\qquad+\emph{YaRN} & \textbf{30.0} & \cellcolor{mygray}\textbf{35.6} & \textbf{30.7} & \cellcolor{mygray}\textbf{38.0} & \textbf{29.6} & \cellcolor{mygray}\textbf{34.1} & \textbf{40.6} & \cellcolor{mygray}\textbf{43.9} & \textbf{24.2} & \cellcolor{mygray}\textbf{32.6} & \textbf{24.1} & \cellcolor{mygray}\textbf{27.8} \\
\midrule
\texttt{Qwen2.5-72B-Instruct} & 39.4 & \cellcolor{mygray}38.8 & \textbf{43.8} & \cellcolor{mygray}42.2 & 36.7 & \cellcolor{mygray}36.7 & 44.4 & \cellcolor{mygray}\textbf{50.0} & 34.0 & \cellcolor{mygray}28.8 & 41.7 & \cellcolor{mygray}39.8 \\
\qquad+\emph{YaRN} & \textbf{42.1} & \cellcolor{mygray}\textbf{43.5} & 42.7 & \cellcolor{mygray}\textbf{47.9} & \textbf{41.8} & \cellcolor{mygray}\textbf{40.8} & \textbf{45.6} & \cellcolor{mygray}48.9 & \textbf{38.1} & \cellcolor{mygray}\textbf{40.9} & \textbf{44.4} & \cellcolor{mygray}\textbf{39.8} \\
\bottomrule
\end{tabular}
}
\caption{Qwen2.5 results (\%) using YaRN on LongBench v2. Higher scores in \textbf{bold}.}
\label{tb:exp_yarn}
\end{table}

\xhdr{Qwen2.5 Results Using YaRN}
Following the guidelines provided in the model card on \url{https://huggingface.co/Qwen/Qwen2.5-72B-Instruct}, we evaluate using YaRN with a scaling factor of 4.0. The results are presented in Table~\ref{tb:exp_yarn}.
YaRN significantly enhances both models' long-context processing ability on LongBench v2, especially on test cases >32k lengths (Medium \& Long).
Additionally, we observe that YaRN has a larger impact on model performance under the CoT setting, though the underlying reasons for this remain unclear.

\xhdr{Compensated Results}
The compensated results that account for invalid outputs are shown in Table~\ref{tb:exp_comp}. We can see that the proportion of invalid outputs is relatively small, and it does not affect the conclusions drawn from our experimental results.

\begin{table}[t]
\centering
\resizebox{\linewidth}{!}{
\begin{tabular}{l|cc|cc|cc|cc|cc|cc|cc}
\toprule
 &  & & & & \multicolumn{4}{|c}{\textbf{Difficulty}} & \multicolumn{6}{|c}{\textbf{Length (<32k; 32k-128k; >128k)}} \\
\cmidrule(r){1-5} \cmidrule(lr){6-9} \cmidrule(l){10-15}
\textbf{Model} & \multicolumn{2}{c|}{\textbf{Overall}} & \multicolumn{2}{c|}{\textbf{Invalid}} & \multicolumn{2}{c|}{\textbf{Easy}} & \multicolumn{2}{c|}{\textbf{Hard}} & \multicolumn{2}{c|}{\textbf{Short}} & \multicolumn{2}{c|}{\textbf{Medium}} & \multicolumn{2}{c}{\textbf{Long}} \\ 
\midrule
\multicolumn{15}{l}{\emph{Open-source models}} \\
\texttt{GLM-4-9B-Chat} & 30.4 & \cellcolor{mygray}32.2 & 0.8 & \cellcolor{mygray}5.6 & 31.1 & \cellcolor{mygray}36.6 & 30.0 & \cellcolor{mygray}29.5 & 34.0 & \cellcolor{mygray}36.2 & 30.0 & \cellcolor{mygray}31.9 & 25.2 & \cellcolor{mygray}26.2 \\
\texttt{Llama-3.1-8B-Instruct} & 31.0 & \cellcolor{mygray}30.5 & 3.8 & \cellcolor{mygray}0.4 & 32.0 & \cellcolor{mygray}36.5 & 30.3 & \cellcolor{mygray}26.8 & 37.6 & \cellcolor{mygray}34.4 & 27.9 & \cellcolor{mygray}31.7 & 25.9 & \cellcolor{mygray}21.5 \\
\texttt{Llama-3.1-70B-Instruct} & 31.7 & \cellcolor{mygray}36.6 & 0.2 & \cellcolor{mygray}1.8 & 32.3 & \cellcolor{mygray}36.3 & 31.3 & \cellcolor{mygray}36.8 & 41.2 & \cellcolor{mygray}45.6 & 27.4 & \cellcolor{mygray}34.1 & 24.1 & \cellcolor{mygray}26.9 \\
\texttt{Llama-3.3-70B-Instruct} & 31.0 & \cellcolor{mygray}36.6 & 4.6 & \cellcolor{mygray}1.8 & 35.8 & \cellcolor{mygray}38.5 & 28.0 & \cellcolor{mygray}35.5 & 39.9 & \cellcolor{mygray}45.6 & 27.0 & \cellcolor{mygray}33.4 & 24.1 & \cellcolor{mygray}28.2 \\
\texttt{Llama-3.1-Nemotron-70B-Instruct} & 31.8 & \cellcolor{mygray}37.2 & 3.2 & \cellcolor{mygray}8.2 & 33.6 & \cellcolor{mygray}39.5 & 30.7 & \cellcolor{mygray}35.9 & 40.4 & \cellcolor{mygray}47.8 & 28.0 & \cellcolor{mygray}32.1 & 25.0 & \cellcolor{mygray}29.9 \\
\texttt{Qwen2.5-7B-Instruct} & 28.9 & \cellcolor{mygray}30.0 & 7.4 & \cellcolor{mygray}0.8 & 31.5 & \cellcolor{mygray}31.0 & 27.3 & \cellcolor{mygray}29.4 & 39.0 & \cellcolor{mygray}35.7 & 25.5 & \cellcolor{mygray}26.7 & 18.8 & \cellcolor{mygray}27.1 \\
\texttt{Qwen2.5-72B-Instruct} & \textbf{40.4} & \cellcolor{mygray}\textbf{39.2} & 4.0 & \cellcolor{mygray}1.6 & \textbf{44.4} & \cellcolor{mygray}43.0 & \textbf{37.9} & \cellcolor{mygray}36.8 & \textbf{46.7} & \cellcolor{mygray}\textbf{50.1} & \textbf{34.2} & \cellcolor{mygray}29.4 & \textbf{42.1} & \cellcolor{mygray}\textbf{40.3} \\
\texttt{Mistral-Large-Instruct-2407} & 30.9 & \cellcolor{mygray}34.5 & 16.9 & \cellcolor{mygray}3.6 & 34.9 & \cellcolor{mygray}35.4 & 28.4 & \cellcolor{mygray}33.9 & 37.8 & \cellcolor{mygray}41.7 & 25.6 & \cellcolor{mygray}31.6 & 29.9 & \cellcolor{mygray}28.2 \\
\texttt{Mistral-Large-Instruct-2411} & 35.7 & \cellcolor{mygray}41.0 & 5.4 & \cellcolor{mygray}5.6 & 40.1 & \cellcolor{mygray}\textbf{45.3} & 33.0 & \cellcolor{mygray}\textbf{38.3} & 43.3 & \cellcolor{mygray}47.9 & 31.7 & \cellcolor{mygray}\textbf{36.0} & 31.0 & \cellcolor{mygray}39.1 \\
\texttt{c4ai-command-r-plus-08-2024} & 28.8 & \cellcolor{mygray}32.0 & 3.8 & \cellcolor{mygray}1.4 & 31.0 & \cellcolor{mygray}34.9 & 27.4 & \cellcolor{mygray}30.1 & 37.4 & \cellcolor{mygray}39.6 & 25.2 & \cellcolor{mygray}24.8 & 21.5 & \cellcolor{mygray}33.6 \\
\midrule
\multicolumn{15}{l}{\emph{Proprietary models}} \\
\texttt{GLM-4-Plus} & 44.6 & \cellcolor{mygray}47.6 & 1.0 & \cellcolor{mygray}5.8 & 47.5 & \cellcolor{mygray}53.5 & 42.8 & \cellcolor{mygray}43.9 & 50.7 & \cellcolor{mygray}54.7 & 46.5 & \cellcolor{mygray}46.2 & 30.6 & \cellcolor{mygray}38.4 \\
\texttt{GPT-4o-mini-2024-07-18} & 29.8 & \cellcolor{mygray}32.6 & 2.0 & \cellcolor{mygray}0.8 & 31.8 & \cellcolor{mygray}32.8 & 28.5 & \cellcolor{mygray}32.5 & 32.5 & \cellcolor{mygray}35.1 & 29.0 & \cellcolor{mygray}31.7 & 26.6 & \cellcolor{mygray}30.1 \\
\texttt{GPT-4o-2024-08-06} & 50.2 & \cellcolor{mygray}51.3 & 0.2 & \cellcolor{mygray}0.4 & 57.4 & \cellcolor{mygray}58.2 & 45.7 & \cellcolor{mygray}47.1 & 53.5 & \cellcolor{mygray}53.9 & 52.4 & \cellcolor{mygray}50.8 & 40.2 & \cellcolor{mygray}47.9 \\
\texttt{gpt-4o-2024-11-20} & 47.4 & \cellcolor{mygray}51.7 & 5.6 & \cellcolor{mygray}1.2 & 52.9 & \cellcolor{mygray}54.7 & 44.0 & \cellcolor{mygray}49.8 & 50.1 & \cellcolor{mygray}60.1 & 48.5 & \cellcolor{mygray}48.7 & 40.7 & \cellcolor{mygray}43.8 \\
\texttt{o1-mini-2024-09-12} & 38.3 & \cellcolor{mygray}39.4 & 1.8 & \cellcolor{mygray}2.0 & 39.7 & \cellcolor{mygray}43.4 & 37.4 & \cellcolor{mygray}36.9 & 48.7 & \cellcolor{mygray}49.6 & 34.0 & \cellcolor{mygray}33.5 & 29.0 & \cellcolor{mygray}34.3 \\
\texttt{o1-preview-2024-09-12} & \textbf{57.9} & \cellcolor{mygray}\textbf{57.1} & 0.8 & \cellcolor{mygray}3.4 & \textbf{67.1} & \cellcolor{mygray}\textbf{60.5} & \textbf{52.3} & \cellcolor{mygray}\textbf{55.0} & \textbf{62.7} & \cellcolor{mygray}\textbf{65.3} & \textbf{53.8} & \cellcolor{mygray}\textbf{51.1} & \textbf{58.3} & \cellcolor{mygray}\textbf{55.5} \\
\texttt{Claude-3.5-Sonnet-20241022} & 44.4 & \cellcolor{mygray}50.4 & 13.9 & \cellcolor{mygray}14.9 & 51.7 & \cellcolor{mygray}59.6 & 40.0 & \cellcolor{mygray}44.8 & 49.2 & \cellcolor{mygray}56.0 & 41.9 & \cellcolor{mygray}46.5 & 41.7 & \cellcolor{mygray}49.1 \\
\midrule
\cellcolor{mypink}\emph{Human} & \multicolumn{2}{c|}{\cellcolor{mypink}55.7} & \multicolumn{2}{c|}{\cellcolor{mypink}8.2} & \multicolumn{2}{c|}{\cellcolor{mypink}100} & \multicolumn{2}{c|}{\cellcolor{mypink}28.4} & \multicolumn{2}{c|}{\cellcolor{mypink}49.3} & \multicolumn{2}{c|}{\cellcolor{mypink}60.3} & \multicolumn{2}{c}{\cellcolor{mypink}57.2} \\
\bottomrule
\end{tabular}
}
\caption{Compensated results (\%) on LongBench v2. Due to the model's occasional refusal to answer or errors in the answer format under our zero-shot prompting, which leads to the failure of parsing selected options, these cases are classified as \emph{invalid} outputs (invalid output rate presented in the table). We account for such cases by applying a 25\% accuracy rate, and the compensated results are shown in this table. We also apply this compensation method to human baselines for cases where the human response is ``I don't know the answer''.}
\label{tb:exp_comp}
\end{table}

\end{document}